\documentclass[journal=jacsat,manuscript=article]{achemso}

\usepackage[version=3]{mhchem} 
\usepackage{siunitx}
\usepackage{booktabs}
\usepackage{subcaption}
\usepackage{amssymb}
\usepackage{microtype}
\usepackage{xr}
\author{Zeno Romero}
\affiliation{Laboratory of Engineering Thermodynamics, RPTU Kaiserslautern, Erwin-Schrödinger-Str. 44, 67663 Kaiserslautern, Germany}
\author{Kerstin Münnemann}
\affiliation{Laboratory of Engineering Thermodynamics, RPTU Kaiserslautern, Erwin-Schrödinger-Str. 44, 67663 Kaiserslautern, Germany}
\author{Hans Hasse}
\affiliation{Laboratory of Engineering Thermodynamics, RPTU Kaiserslautern, Erwin-Schrödinger-Str. 44, 67663 Kaiserslautern, Germany}
\author{Fabian Jirasek}
\affiliation{Laboratory of Engineering Thermodynamics, RPTU Kaiserslautern, Erwin-Schrödinger-Str. 44, 67663 Kaiserslautern, Germany}
\email{fabian.jirasek@rptu.de}

\title[]
  {Prediction of Diffusion Coefficients in Mixtures with Tensor Completion}

\abbreviations{}
\keywords{}

\begin{document}







\begin{abstract}

Predicting diffusion coefficients in mixtures is crucial for many applications, as experimental data remain scarce, and machine learning (ML) offers promising alternatives to established semi-empirical models. Among ML models, matrix completion methods (MCMs) have proven effective in predicting thermophysical properties, including diffusion coefficients in binary mixtures. However, MCMs are restricted to single-temperature predictions, and their accuracy depends strongly on the availability of high-quality experimental data for each temperature of interest. In this work, we address this challenge by presenting a hybrid tensor completion method (TCM) for predicting temperature-dependent diffusion coefficients at infinite dilution in binary mixtures. The TCM employs a Tucker decomposition and is jointly trained on experimental data for diffusion coefficients at infinite dilution in binary systems at 298~K, 313~K, and 333~K. Predictions from the semi-empirical SEGWE model serve as prior knowledge within a Bayesian training framework. The TCM then extrapolates linearly to any temperature between 268~K and 378~K, achieving markedly improved prediction accuracy compared to established models across all studied temperatures. To further enhance predictive performance, the experimental database was expanded using active learning (AL) strategies for targeted acquisition of new diffusion data by pulsed-field gradient (PFG) NMR measurements. Diffusion coefficients at infinite dilution in 19 solute + solvent systems were measured at 298~K, 313~K, and 333~K. Incorporating these results yields a substantial improvement in the TCM's predictive accuracy. These findings highlight the potential of combining data-efficient ML methods with adaptive experimentation to advance predictive modeling of transport properties.

\end{abstract}

\section{Introduction}

Diffusion is the fundamental process governing mass transport and determines the rate and selectivity of many processes in nature and technology. Despite their importance, experimental data on diffusion coefficients remain scarce, as there are simply too many mixtures of interest to study more than a tiny fraction of them by tedious diffusion measurements. Accordingly, the development of reliable prediction methods for diffusion coefficients is an important research topic\cite{PGL,Evans2018,GuevaraCarrion2016,Parez2013,Schmitt2024b,Kankanamge2025,Bellaire2022,Lenahan2024}, which is currently evolving rapidly, driven by machine learning (ML)\cite{Grossmann2022,Jirasek2023,Aniceto2024}. We focus here on the demanding, practically important task of predicting diffusion coefficients in liquid mixtures.

There are two distinct classes of diffusion coefficients: \emph{mutual} diffusion coefficients, which describe the motion of collectives of molecules driven by chemical-potential gradients, and \emph{self}-diffusion coefficients, which characterize the Brownian motion of individual molecules \cite{Taylor1993}. Liquid-phase mutual diffusion is commonly modeled using either the Maxwell–Stefan or Fickian framework. Established measurement methods include diaphragm cells \cite{Tham1967}, Taylor-dispersion experiments \cite{Taylor1953}, dynamic light scattering \cite{Mountain1969}, and concentration-profile monitoring in quiescent fluids \cite{Bellaire2022b}. 

Pulsed-field gradient NMR allows a calibration-free and accurate self-diffusion measurement in both pure liquids and liquid mixtures \cite{Bellaire2020, Bellaire2022, Specht2023, Mross2024}. It uses a short magnetic-field-gradient pulse to label nuclear spins with position-dependent phases, followed by a second gradient pulse, applied after a defined delay, which rephases the spins. Molecular diffusion during this delay leads to incomplete rephasing, resulting in a decrease in signal intensity that scales with the diffusion coefficient; specifically, the faster the diffusion, the greater the signal reduction.

The diffusion coefficient $D_{ij}^\infty$ of a solute $i$ at infinite dilution in a solvent $j$ is of particular interest for several reasons: at this limit, self- and mutual diffusion coefficients coincide, and the Maxwell–Stefan and Fickian descriptions become identical. Moreover, if both infinite-dilution coefficients in a binary mixture are known ($i$ in $j$ and $j$ in $i$), an extrapolation to finite concentrations is possible, e.g., via the empirical Vignes correlation \cite{Vignes1966}, with possible extension to multi-component systems \cite{Taylor1993}. The semi-empirical Stokes-Einstein–Gierer–Wirtz estimation (SEGWE) model \cite{Evans2018} is currently the most accurate semi-empirical model for the prediction of $D_{ij}^\infty$.


Recently, we have introduced matrix completion methods (MCMs) from ML, which are well established in recommender systems~\cite {Koren2009}, for predicting $D_{ij}^\infty$ at 298 K~\cite{Grossmann2022}. The key idea is to represent experimental data measured for different binary mixtures as a matrix whose rows and columns correspond to components $i$ and $j$, with each entry containing the available data for mixture $i+j$ \cite{Jirasek2020, Jirasek2021}. Because this matrix is sparsely populated with experimental data, predicting the properties of unstudied mixtures reduces to a matrix completion problem. MCMs have since been developed for various thermodynamic properties, including activity coefficients \cite{Jirasek2020, Jirasek2020b, Damay2021, Gond2025, Hayer2025cosmo, Zenn2025}, Henry’s-law constants \cite{Hayer2022, Hayer2024}, and diffusion coefficients \cite{Grossmann2022,Romero2025}, as well as for pair-interaction parameters in thermodynamic models \cite{Jirasek2022, Hayer2025, Hoffmann2024, Jirasek2023b, Jirasek2023c}.

For predicting $D_{ij}^\infty$, hybrid approaches that incorporate prior physical knowledge from the SEGWE model \cite{Evans2018} in the MCM training are especially promising, outperforming all available semi-empirical alternatives in prediction accuracy at 298 K~\cite{Grossmann2022}. However, because MCMs require a matrix structure in their training data, they are restricted to predicting a single property of binary mixtures under fixed conditions; for $D_{ij}^\infty$, this means single-temperature predictions. Industrial practice, however, demands knowledge across a wide temperature range, not only at 298~K, where data are even sparser. 

There are two general ways for extending MCMs to higher dimensions, e.g., for predicting temperature-dependent thermodynamic properties. The first route is feasible if the temperature dependence of the property of interest is known. Then, the MCMs can be applied to predict the mixture-specific parameters of the equation describing the temperature dependence. This route was introduced by Damay et al.~\cite{Damay2021} for predicting temperature-dependent activity coefficients at infinite dilution using the Gibbs-Helmholtz relation. The second route is to extend the MCM to a tensor completion method (TCM), whereby a three-dimensional tensor is spanned by the two components that make up the mixtures and the temperature. The TCM concept was transferred to thermodynamics by Damay et al.~\cite{Damay2023}, who again considered the temperature-dependent prediction of activity coefficients at infinite dilution. 

The TCM approach, unlike the first route, is also applicable when no general equation describing the temperature dependence is available. Liquid-phase diffusion coefficients are sometimes approximated as having a linear temperature dependence, consistent with the Stokes-Einstein theory\cite{Sutherland1905} if the temperature dependence of the solvent viscosity is neglected. However, this is not generally applicable, making TCMs an interesting option in this field.

The predictive capabilities of an ML model can also be enhanced by purposefully incorporating new data into the training set through active learning (AL) methods. Such AL strategies iteratively select the presumed most informative data points for experimental measurement, without prior knowledge of their values, and thereby aim to maximize model improvement with minimal experimental effort. To this end, a query strategy is employed within an AL framework~\cite{Settles2009}. In a previous work, we found that uncertainty sampling, i.e., selecting the data point to be measured based on the current model's largest uncertainty, is an effective query strategy for improving the performance of an MCM for the prediction of $D_{ij}^\infty$.\cite{Romero2025}

In this work, we present a novel hybrid TCM for predicting $D_{ij}^\infty$ across temperatures. Our method employs a Tucker decomposition \cite{Tucker1963}, analogous to that of Damay et al. \cite{Damay2023}, and integrates SEGWE priors, following our earlier MCM approach \cite{Grossmann2022}. This TCM is trained on $D_{ij}^\infty$ data at 298~K, 313~K, and 333~K. While $D_{ij}^\infty$ are obviously of interest well beyond these three discrete temperatures, substantially less experimental information is available outside this range, preventing the development of MCMs for predicting $D_{ij}^\infty$ at these temperatures. However, the developed TCM can also predict temperatures absent from the training set, and we evaluate its predictions at temperatures between 268~K and 378~K, comparing them to experimental diffusion data and SEGWE\cite{Evans2018} predictions within this extended range.

Furthermore, we extend the available experimental data on $D_{ij}^\infty$ at 298~K, 313~K, and 333~K by measuring $D_{ij}^\infty$ by Pulsed-Field Gradient NMR spectroscopy \cite{Bellaire2020,Bellaire2022,Mross2024} and selecting the measured systems using AL~\cite{Settles2009,Romero2025} and uncertainty sampling. We systematically evaluate the influence of the new training data on the prediction accuracy.

\section{Methodology}
\subsection{Database}

Experimental data for liquid-phase $D_{ij}^\infty$ at 298~K in binary mixtures were, on the one hand, taken from the database of Großmann \textit{et al.} \cite{Grossmann2022}, which is based on data from the Dortmund Data Bank 2019 \cite{ddb} and several other sources. Most of the data for $D_{ij}^\infty$ reported by Großmann \textit{et al.}\cite{Grossmann2022} were obtained from an extrapolation of data at finite concentrations. After extending the database of Großmann et al.\cite{Grossmann2022} with new $D_{ij}^\infty$ from the 2025 version of the DDB \cite{ddb} in this work and data from our previous works\cite{Romero2025,Mross2024}, the three temperatures for which by far the most data were available are 298~K, 313~K, and 333~K (cf. Figure S1 in the Supporting Information). Thus, we carried out a comprehensive literature search for $D_{ij}^\infty$ at those temperatures, adding 89 additional data points at 313~K and 333~K from various sources \cite{ddb,tominaga1984limiting,tominaga1990diffusion,Tominaga1990,Hsieh2012,Li1990,Schatzberg1965,Lin2008, hurle1982tracer,tyn1975temperature,sanni1971diffusion,bonoli1968diffusion,te1995diffusion,alizadeh1982mutual,anderson1958mutual,hashim2007diffusion,yumet1985tracer,chen1983tracer,clark1986mutual,Mross2024,Wagner2025,Romero2025}. In all cases, we used the extrapolation scheme of Grossmann et al.\cite{Grossmann2022}  to obtain the $D_{ij}^\infty$ if the data were not directly reported at infinite dilution in the literature. To facilitate model evaluation via leave-one-out analysis, data were filtered to include only solutes $i$ and solvents $j$ for which experimental $D_{ij}^\infty$ data points were available in at least two different mixtures $i+j$. 

\begin{figure}[H]
\centering
\includegraphics[width=1\textwidth]{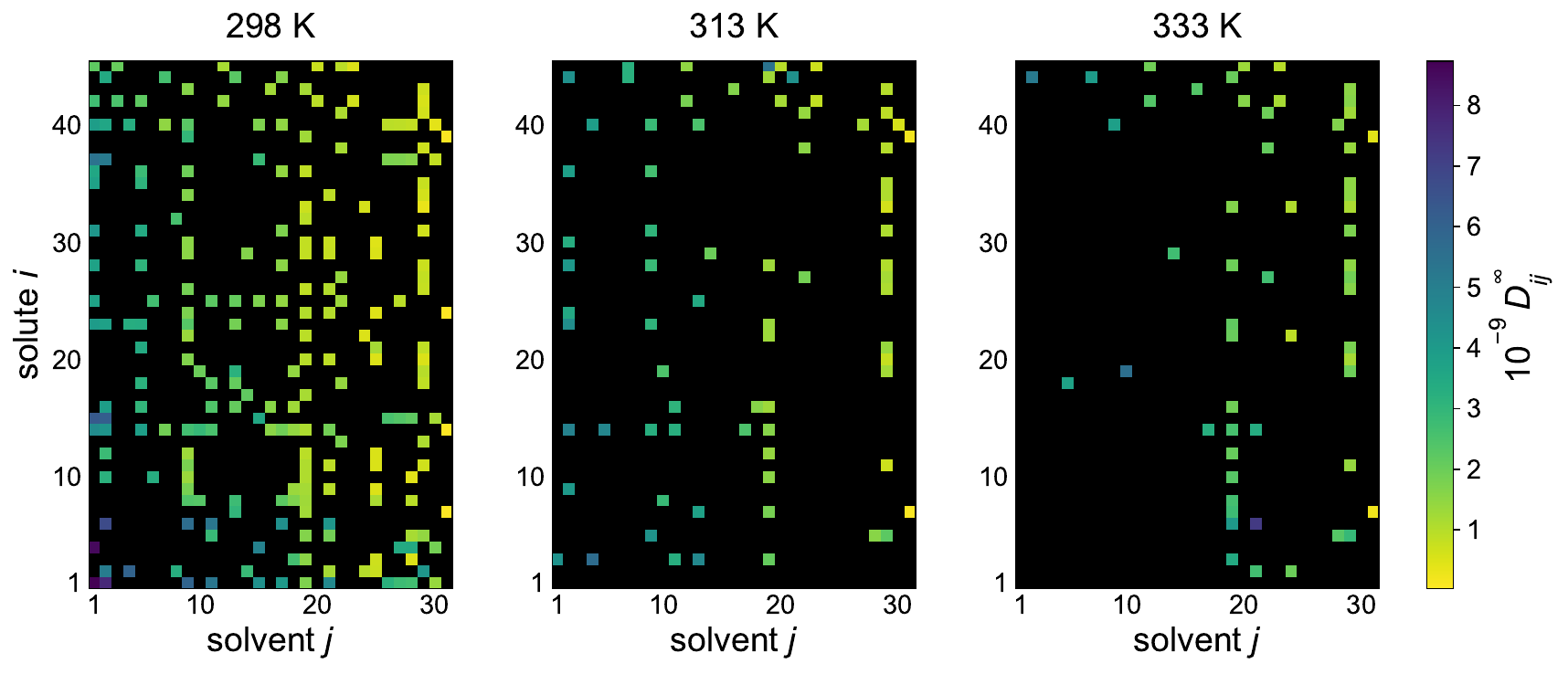}
\caption{Experimental data for liquid-phase diffusion coefficients $D_{ij}^{\infty}$ of solutes $i$ at infinite dilution in solvents $j$ at temperatures 298~K, 313~K, and 333~K in the database used as the starting point in the present work. Numbers identify solutes and solvents, cf. Tables S1 and S2 in the Supporting Information. Solutes are ordered with respect to their molar mass (bottom: low; top: high), and solvents are ordered with respect to their viscosity (left: low; right: high). The color code indicates the value of $D_{ij}^{\infty\text{,exp}}$, and black cells denote missing data.}
\label{fig:db}
\end{figure}

The resulting database of experimental values of $D_{ij}^\infty$ used in this work consists of 224 data points at 298~±~1~K, 75 data points at 313~$\pm$~1~K, and 56 data points at 333~$\pm$~1~K. It covers 45 different solutes $i$ infinitely diluted in 31 different solvents $j$. The data can be arranged in temperature-specific matrices, where the rows represent the solutes and the columns represent the solvents, cf. Figure \ref{fig:db}. The solutes and solvents included in the matrix are listed in Tables S1 and S2 in the Supporting Information. The included compounds consist mostly of water and organic molecules that are liquid at ambient conditions and have low reactivity. The solutes additionally contain 5 substances that are gaseous at ambient conditions. Values of $D_{ij}^{\infty}$ range from $10^{-11}$ to $10^{-8}$ \SI{}{\meter\squared\per\second}.

The dataset can also be represented as a third-order tensor, with the three dimensions being the solutes, solvents, and temperatures. In total, this tensor has 4185 elements, of which, however, only 8.5~\% are occupied by experimental data for $D_{ij}^{\infty}$. As shown in Figure \ref{fig:db}, this tensor is not only sparsely but also heterogeneously occupied. Most data are available at 298~K, and there are some solvents and solutes for which much more data are available than for the others; there are even several solvents and solutes for which no data are available at 313~K and 333~K at all. Details on the data availability per temperature are listed in Table \ref{tab:dbinfo}.

\begin{table}
    \centering
    \begin{tabular}{lccc}
         \toprule
         $T$  & 298 K & 313 K & 333 K \\
         \midrule
         Number of data points & 224 & 75 & 56 \\
         Matrix occupation rate & $16.1 \: \%$ & $5.4\: \% $  & $4.0 \: \%$ \\
         Number of available solvents & 31 & 24 & 18 \\
         Number of available solutes & 45 & 35 & 33 \\
         \bottomrule
    \end{tabular}
    \caption{Information on the availability of $D_{ij}^{\infty}$ in our database for each studied temperature, cf. Figure \ref{fig:db}.}
    \label{tab:dbinfo}
\end{table}

\begin{figure}[H]
\centering
\includegraphics[width=1\textwidth]{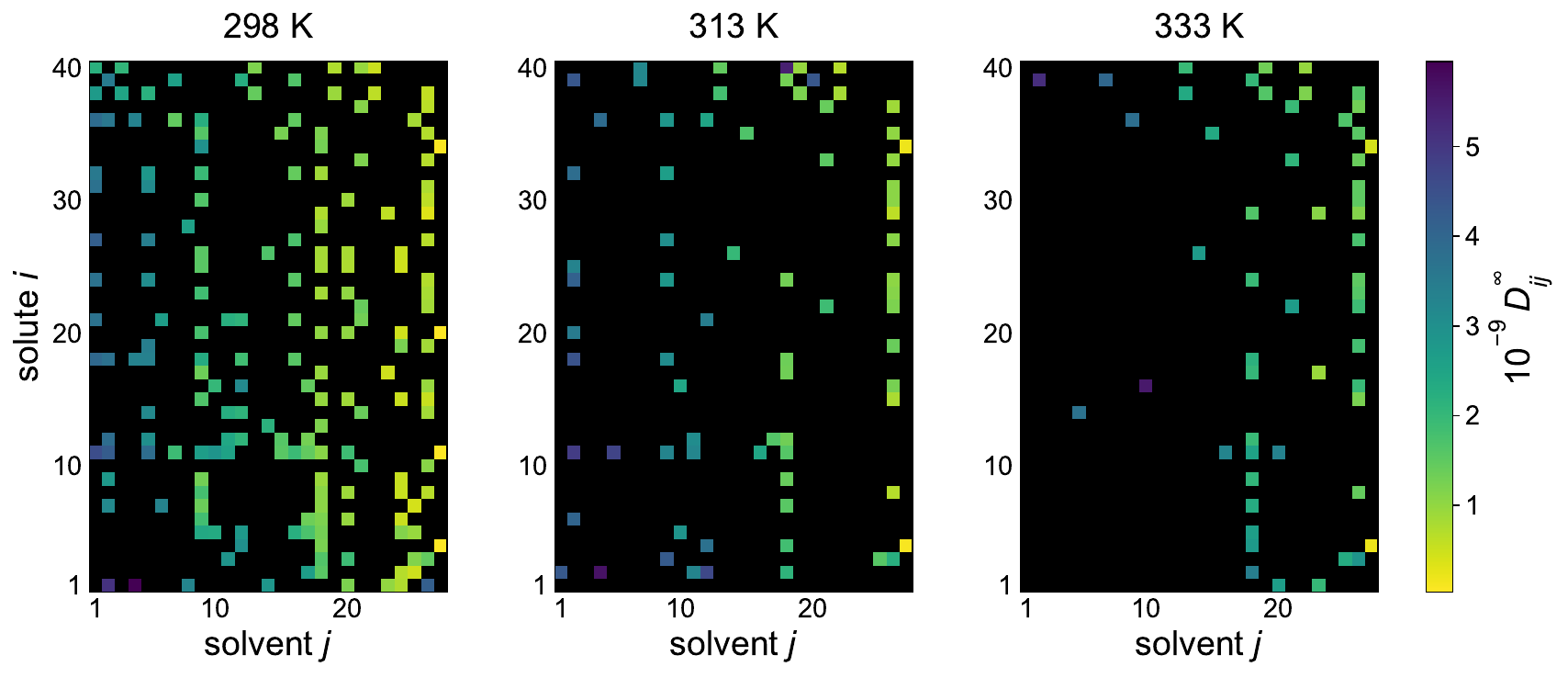}
\caption{Experimental data for liquid-phase diffusion coefficients $D_{ij}^{\infty\text{,exp}}$ of solutes $i$ at infinite dilution in solvents $j$ at temperatures 298~K, 313~K, and 333~K in the database used as starting point for the AL study. Numbers identify solutes and solvents, cf. Tables S3 and S4 in the Supporting Information. Solutes are ordered with respect to their molar mass (bottom: low; top: high), and solvents are ordered with respect to their viscosity (left: low; right: high). The color code indicates the value of $D_{ij}^{\infty}$, and black cells denote missing data.}
\label{fig:db_al}
\end{figure}

To facilitate sample handling during the measurements planned by AL in this work, we further filtered the data from Figure \ref{fig:db} to yield a reduced dataset by excluding all gaseous compounds at ambient conditions. Due to this exclusion, we again had to filter data so only solutes $i$ and solvents $j$ for which experimental data points for $D^{\infty}_{ij}(T)$ in at least two different mixtures $i+j$ were available were included. We chose to exclude these substances beforehand to follow the query strategy as closely as possible, rather than intervening during the AL workflow by skipping selected systems. The temperature-specific matrix arrangement of this reduced database is shown in Figure \ref{fig:db_al} and is the underlying database for the experimental AL workflow. The solutes and solvents included in this matrix are listed in Tables S3 and S4 in the Supporting Information.

While we focus on three temperatures here to compare TCM predictions with temperature-specific MCMs, a continuous-temperature approach, as explained in the following chapters, enables us to generalize to arbitrary temperatures. To assess the performance of the TCM over the continuous temperature range, we use another dataset from the DDB 2025 \cite{ddb} spanning 268~K to 378~K (but excluding 298~K, 313~K, and 333~K) and containing 98 data points. No data for these temperatures were, however, used for training the TCM.

In addition to the experimental data, henceforth called $D_{ij}^{\infty\mathrm{,exp}}$, a synthetic database, $D_{ij}^{\infty\mathrm{,SEGWE}}$, was used for pre-training both temperature-specific MCMs and the TCM. This synthetic database consists of predictions of $D_{ij}^{\infty}$ at 298~K, 313~K, and 333~K using the SEGWE model \cite{Evans2018} for the same solutes and solvents as in the experimental database. The solvent viscosities required for the SEGWE model \cite{Evans2018} were calculated using the correlations and parameters from the DIPPR database \cite{DIPPR2024}, and the effective density (a parameter in the SEGWE model) was set to the recommended value $\rho_{\text{eff}}=627$ kg m\textsuperscript{-3} \cite{Evans2018}, as it was done in our previous works~\cite{Grossmann2022,Romero2025}.

\subsection{Matrix Completion Method}

In the present work, we have used a hybrid MCM that combines experimental data on $D_{ij}^\infty$ with synthetic data obtained with the SEGWE model during the training in a Bayesian framework and was introduced and termed "MCM-Whisky" in our previous works\cite{Jirasek2020b,Grossmann2022,Romero2025} to predict isothermal $D_{ij}^\infty$ as a benchmark for comparison with the newly developed TCM, which predicts temperature-dependent $D_{ij}^\infty$. This MCM is based on a low-rank matrix factorization of an isothermal $D_{ij}^\infty$ matrix, which is only sparsely populated with experimental data (cf. Figure \ref{fig:db}). The training of the MCM involves two steps with different (experimental or synthetic) training data; in both steps, the training data are modeled as:
\begin{equation}
\label{eqn:mcm}
\ln D_{ij}^\infty = u_i \cdot v_j + \varepsilon_{ij}
\end{equation}
where $u_i$ and $v_j$ are component-specific feature vectors of the solute $i$ and the solvent $j$, representing the fitting parameters of the model, and $\varepsilon_{ij}$ are the deviations between model predictions and the training data. Both feature vectors have length $K=2$, which is a hyperparameter of the model and was adapted from our previous work\cite{Grossmann2022,Romero2025}.

In the first training step, an MCM is trained on the complete synthetic $\ln D_{ij}^{\infty\text{,SEGWE}}$ data matrix according to equation~(\ref{eqn:mcm}) using uninformed normal prior distributions with $\mu_0 = 0$ and $\sigma_0=1$ and a Cauchy likelihood with scale parameter $\lambda=0.2$. The resulting preliminary features $u_i^*$ and $v_j^*$, obtained by minimizing the residuals $\varepsilon_{ij}$ and described by the posterior probability distributions from this first training step, were scaled and then used as informed normal prior distributions for the second MCM trained on the sparse $\ln D_{ij}^{\infty\text{,exp}}$ matrix following equation~(\ref{eqn:mcm}), again using a Cauchy likelihood with scale parameter $\lambda=0.2$, minimizing the residuals $\varepsilon_{ij}$ now referring to the experimental data, and resulting in the final solute and solvent features, $u_i$ and $v_j$. For the scaling, the mean of the posterior distributions of $u_i^*$ and $v_j^*$ was adopted, whereas their standard deviation was scaled with a constant factor to obtain an average value (averaged over all solutes $i$ and solvents $j$) of $\bar{\sigma} = 0.5$. The resulting distributions were finally multiplied by the uninformed normal prior ($\mu_0 = 0$, $\sigma_0=1$) used in the first training step. This probabilistic hybrid approach allows prior physical information from the SEGWE model to be incorporated into the MCM, while maintaining the flexibility of the model to adapt to experimental data. More details on this hybrid approach can be found in our earlier work~\cite{Jirasek2020b,Grossmann2022,Romero2025}.

Since we use a Bayesian approach for this second training step as well, we obtain posterior distributions over the model parameters after training, from which probability distributions for each predicted matrix entry can be calculated using equation~(\ref{eqn:mcm}). The mean of these distributions was considered as the predicted diffusion coefficient $\ln D_{ij}^{\infty\text{,pred}}$. Furthermore, from the obtained probability distributions, the standard deviation $\sigma_{ij}$ was calculated as a measure for model uncertainty.

The MCM approach was used to predict isothermal diffusion coefficients. Hence, an individual MCM was trained on data for a single temperature, i.e., a single matrix from Figure \ref{fig:db}, and generated predictions for the missing values at that same temperature, which was done here for 298, 313, and 333~K. It does not use information on the $D_{ij}^\infty$ at multiple temperatures and cannot extrapolate from one temperature to another. In the following, we will refer to the hybrid MCM approach simply as MCM.

\subsection{Tensor Completion Method}

In this work, we introduce a novel hybrid tensor completion method for the temperature-dependent prediction of $D_{ij}^\infty$ as an extension of the hybrid temperature-specific MCM. For this purpose, we use a low-rank Tucker decomposition modeling the training data as:
\begin{equation}
\label{eqn:tcm}
\ln D_{ij}^\infty(T) =  \sum_{\alpha=1}^{r_u} \sum_{\beta=1}^{r_v} \sum_{\gamma=1}^{r_w} u_{i\alpha} \cdot v_{j\beta} \cdot w_{\gamma}(T) \cdot \kappa_{\alpha\beta\gamma} + \varepsilon_{ijk}
\end{equation}

where $u$, $v$, and $w$ are the latent features of the solute, the solvent, and the temperature, respectively, $r_u$, $r_v$, and $r_w$ are their respective latent feature dimensions, and $\kappa$ is the core tensor, which is used to combine the latent feature matrices into a single tensor. We note that $u$ and $v$ are temperature-independent parameters, whereas $w$ is temperature-dependent.

The TCM approach is a priori discrete with respect to temperatures and was trained and evaluated simultaneously at the three temperatures 298~K, 313~K, and 333~K. For each temperature, it learns an independent set of latent temperature features $w_{\gamma}(T)$, which contain no prior information about temperature and do not incorporate any physically motivated scaling. However, these learned features $w_{\gamma}(T)$ were subsequently correlated with the temperature $T$ to enable their prediction at any temperature, not just the discrete ones. The correlation of the temperature features and the application of the TCM for predictions across a broad temperature range is discussed in the Results section.



\begin{figure}[H]
    \centering
    \includegraphics[width=0.45\textwidth]{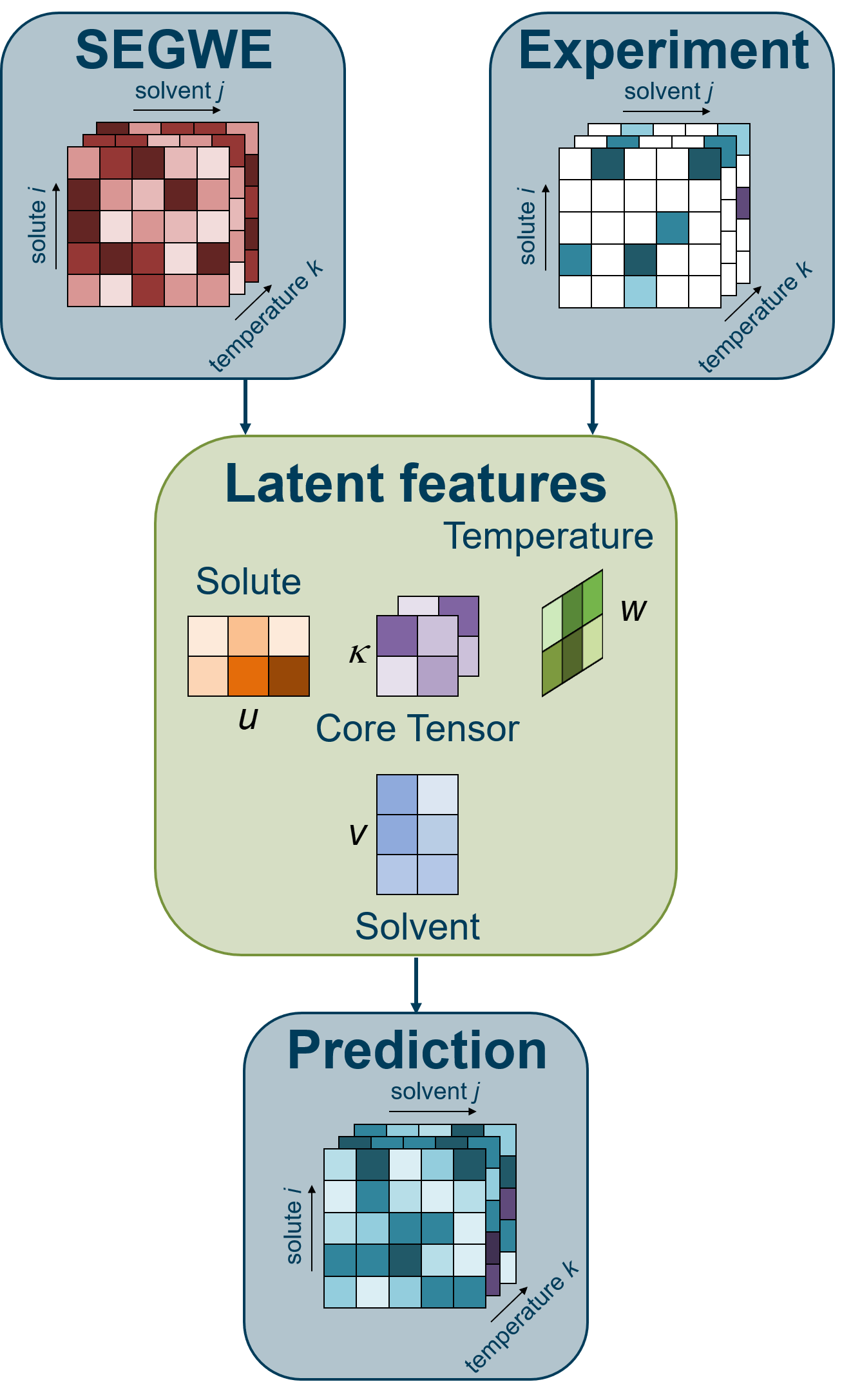}
       \caption{Schematic representation of the hybrid TCM for predicting temperature-dependent $D_{ij}^\infty$ developed in this work. The TCM incorporates prior information from the SEGWE model \cite{Evans2018} and uses the Tucker decomposition for tensor factorization.}
    \label{fig:TCM}
\end{figure}

Tucker decomposition, which becomes equivalent to canonical polyadic decomposition if $\kappa$ is the unit tensor, was chosen because of its flexibility by introducing $\kappa$. Analogous to the MCM, we propose the hybrid TCM approach, which additionally incorporates SEGWE\cite{Evans2018} predictions into its training. This approach consists of two steps, schematically shown in Figure \ref{fig:TCM}.

Analogously to MCM, the TCM is trained in two steps. First, the TCM is fitted to the fully completed synthetic tensor of $\ln D_{ij}^{\infty,\text{SEGWE}}$ using uninformed normal prior distributions with $\mu_0 = 0$ and $\sigma_0=1$ and a Cauchy likelihood with scale parameter $\lambda=0.2$. The posterior distributions of the latent features $u^*$, $v^*$, $w^*$, and $\kappa^*$ from this run then serve as priors for a second MCM on the sparse experimental tensor $\ln D_{ij}^{\infty,\text{exp}}$ using a Cauchy likelihood with scale parameter $\lambda=0.2$. For each feature, we keep the posterior mean and rescale its standard deviation by a constant so that the overall average becomes $\bar{\sigma}=0.5$ (averaged over all $i,j,T$). These informed priors are multiplied by the default uninformative prior ($\mu_0=0$, $\sigma_0=1$). Following our earlier work \cite{Grossmann2022,Romero2025}, this scheme injects prior physical knowledge from the SEGWE \cite{Evans2018} model into the TCM while retaining flexibility to fit the experimental data. 

$\ln D_{ij}^{\infty,\text{pred}}$ are calculated analogously to MCM, from the posterior distributions of the model parameters, according to equation (\ref{eqn:tcm}). The mean of the resulting distribution for each tensor entry is taken as $\ln D_{ij}^{\infty,\text{pred}}$, whereas their standard deviation $\sigma_{ij}(T)$ serves as a measure for model uncertainty.

The use of $\kappa$ generally allows different latent feature dimensions $r_u$, $r_v$, and $r_w$, which are the hyperparameters of the model. We have carried out hyperparameter optimization in this work, using system-wise leave-one-out cross-validation; cf. below for details. The results of the hyperparameter study are given in Figure S2 of the Supporting Information. We found that the best prediction accuracy was achieved using $r_u=r_v=r_w=2$. 


\subsection{Active Learning}

The objective of AL is to enhance the predictive capabilities of an ML model by purposefully incorporating new data into the training set. Ideally, these data points are selected to maximize the model's performance gain without prior knowledge of their values. To this end, a query strategy is employed within an AL framework.\cite{Settles2009}

In this work, the ML model to be improved is the TCM for predicting $D^{\infty}_{ij}$ in binary mixtures at 298~K, 313~K, and 333~K. For the AL, we thereby constrain the newly measured data and evaluations to the three discrete temperatures, i.e., the possible solute-solvent-temperature tuples $(i,j,T)$, thereby limiting the experimental space, which would otherwise be infinitely large due to continuous temperature. Consequently, we used a pool-based sampling approach, where all solute-solvent pairs $(i,j)$ for which no $D^{\infty\text{,exp}}_{ij}$ exist at any temperature $T$ comprise the sampling pool $\mathcal{U}$, which contains the solute-solvent pairs from which the query strategy may choose new mixtures to be measured. 

\begin{figure}[H]
    \centering
    \includegraphics[width=0.6\textwidth]{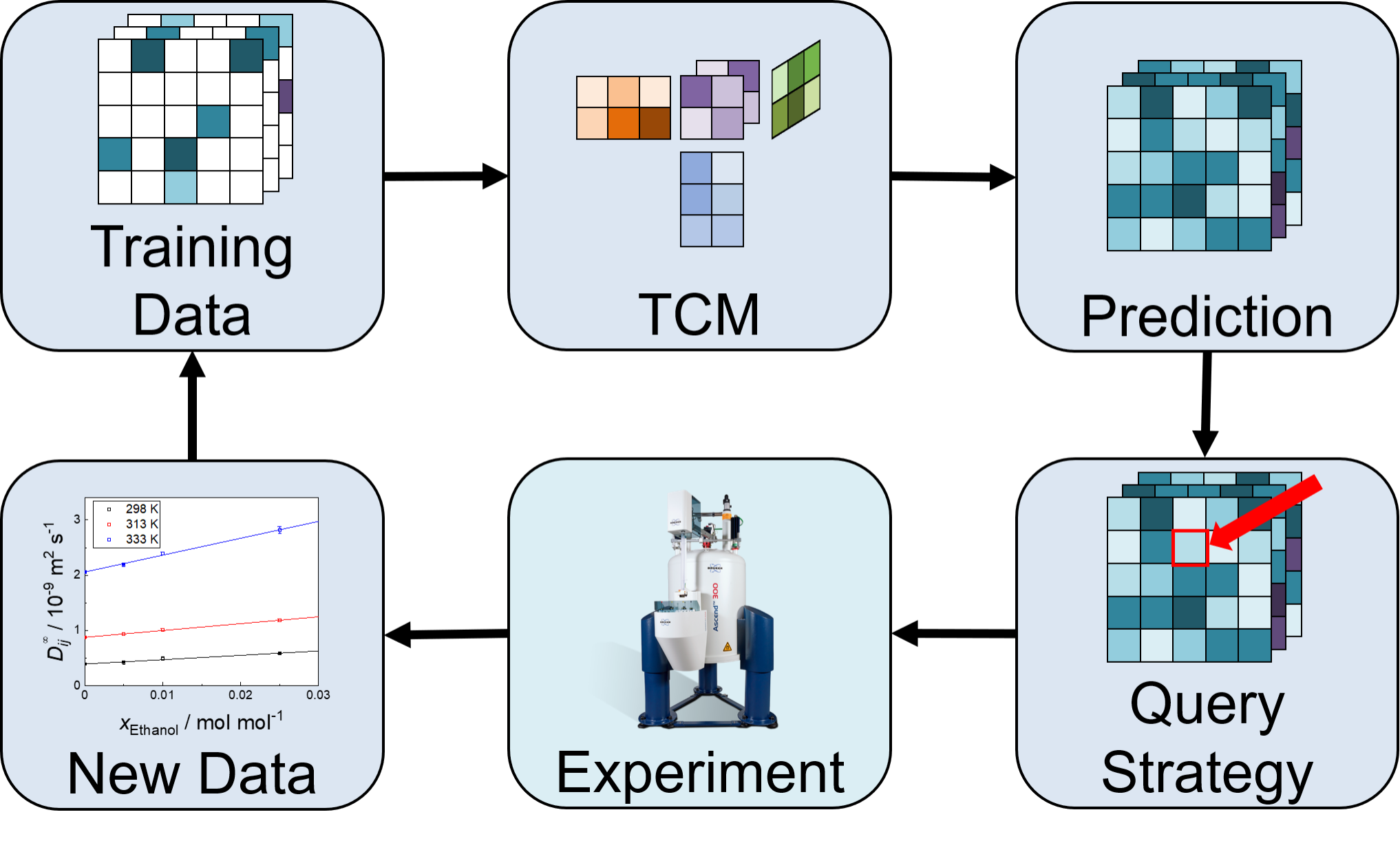}
    \caption{Active learning workflow for the targeted improvement of the TCM developed in this work.}
    \label{fig:AL}
\end{figure}

Figure \ref{fig:AL} shows the general AL framework used in this work, which was adopted from our previous work.\cite{Romero2025}

As illustrated in Figure \ref{fig:AL}, the AL workflow is an iterative process. We begin with the initial training data set, i.e., the initially available experimental data for $D^{\infty}_{ij}$, cf. Figure \ref{fig:db}. The TCM is trained on this data set and can then be used to generate a complete tensor of predicted diffusion coefficients $D^{\infty\text{,pred}}_{ij}$. Based on the obtained predictions, a query strategy is used to select a solute-solvent pair $(i,j)^*\in\mathcal{U}$ for which no experimental data are available at any temperature $T \in \Theta$, where  $\Theta=$~\{\SI{298}{K},~\SI{313}{K},~\SI{333}{K}\}. $D^{\infty\text{,exp}}_{ij}\: \forall \: T\in\Theta$ are then measured for this selected system by PFG NMR spectroscopy at all temperatures $k$, and the new data is subsequently added to the training data set. This procedure is repeated several times, increasing the training data size at each iteration and thus (hopefully) improving the prediction accuracy of the model. The key to this improvement lies in choosing a suitable query strategy.

In our previous study, we found that uncertainty was the most suitable query strategy for improving the prediction of diffusion coefficients with MCMs \cite{Romero2025}, which is why we again use this strategy in this work for the TCM approach. For this purpose, we average the prediction uncertainty $\sigma_{ij}(T)$ resulting from our TCM over the three studied temperatures $T$. This results in a solute-solvent matrix of temperature-averaged uncertainties, from which the entry $(i,j)^*$ with the highest associated prediction uncertainty $\bar{\sigma}_{ij}$ was selected, cf. equation~(\ref{eqn:al_us}), and the new diffusion coefficients $D^{\infty\text{,exp}}_{ij}$ are measured via PFG NMR spectroscopy.
\begin{equation}
\label{eqn:al_us}
(i,j)^*=\underset{(i,j)}{\operatorname{argmax}} \: \frac{1}{|\Theta|} \sum_{T\in\Theta} \sigma_{ij}(T) 
\end{equation}
In practice, uncertainty sampling tends to sample outliers that are not representative of the underlying data distribution, which we also observed in our prior work\cite{Romero2025}. While sampling some outliers can improve the model's prediction accuracy, continuously sampling them yields little new information and leads to redundancy \cite{Kim2024,Zhu2010}. Specifically, in the context of Bayesian MCM (and TCM), after repeated sampling within a single row or column, i.e., repeated sampling of the same solute $i$ or solvent $j$, the information gain by inclusion of another data point in the same row or column is small. At the same time, the posterior for all other compounds can remain wide.\cite{houlsby14}  We thus introduce the simple rule of removing a compound from the sampling pool after it has been sampled in too many consecutive rounds. This approach encourages exploration of the chemical space and reduces redundancy in a simple way. 

\subsection{Computational Details and Evaluation}

Bayesian inference was performed using automatic differentiation variational inference \cite{Kucukelbir2017,Blei2017} implemented in the probabilistic programming language Stan \cite{stan} in its Python package \texttt{CmdStanPy}. The code is provided in the Supporting Information. 

The predictive performance of the models was evaluated using leave-one-out analysis \cite{Cawley2003}. Each model was trained on a subset of the $D^{\infty\text{,exp}}_{ij}$, which includes all available experimental data except for one binary system to be predicted. In the case of the MCM, this means the training data included all experimental data for one specific temperature $T$, except for one solute-solvent pair $(i,j)$, which was then predicted at that same temperature. In the case of the TCM, the training data included all experimental data for all three temperatures $T\in\Theta$, except for one solute-solvent pair $(i,j)$, which was excluded at all temperatures, and predicted at all temperatures. Thus, in all cases, the predictions were made on diffusion coefficients for truly unseen solute-solvent pairs $(i,j)$.

To evaluate the prediction accuracy at each temperature $T\in\Theta$, we computed the absolute relative error ($\mathrm{ARE}_{ij}(T)$) for each data point. These per-point errors were calculated using:
\begin{equation}
\label{eqn:rAE}
\mathrm{ARE}_{ij}(T) =  \left| \frac{D^{\infty\text{,pred}}_{ij}(T) - D^{\infty\text{,exp}}_{ij}(T)}{D^{\infty\text{,exp}}_{ij}(T)} \right|
\end{equation}

These errors were aggregated over the set $\mathcal{L}(T)$, which contains all $(i,j)$ pairs where experimental data are available at temperature $T$ and reported as box plots. The temperature-specific relative mean absolute error ($\mathrm{rMAE}(T)$, cf. equation (\ref{eqn:MAE})) and the relative mean squared error ($\mathrm{rMSE}(T)$, cf. equation (\ref{eqn:MSE})) are also reported: \begin{align}
    \label{eqn:MAE}
    &\mathrm{rMAE}(T) = \frac{1}{|\mathcal{L}(T)|} \sum_{(i,j) \in \mathcal{L}(T)} \left| \frac{D^{\infty\text{,pred}}_{ij}(T) - D^{\infty\text{,exp}}_{ij}(T)}{D^{\infty\text{,exp}}_{ij}(T)} \right|   \\
    \label{eqn:MSE}
    &\mathrm{rMSE}(T) = \frac{1}{|\mathcal{L}(T)|} \sum_{(i,j)\in \mathcal{L}(T)} \left( \frac{D^{\infty\text{,pred}}_{ij}(T) - D^{\infty\text{,exp}}_{ij}(T)}{D^{\infty\text{,exp}}_{ij}(T)} \right)^2
\end{align}

Furthermore, to demonstrate the generalization of the TCM to continuous temperature values, we trained an TCM on all data for $\Theta=$~\{\SI{298}{K},~\SI{313}{K},~\SI{333}{K}\}, and report the errors across the temperature range [\SI{268}{K},~\SI{453}{K}] (excluding data at $T\in\Theta$) using a box plot with aggregated 10~K temperature bins. In all cases, we compare the TCM results to SEGWE predictions and, for $T\in\Theta$, also to isothermal MCMs, using the same error metrics.

\subsection{Measurement of Diffusion Coefficients by PFG NMR Spectroscopy}

In this work, self-diffusion coefficients were measured using PFG NMR, following the method described in our previous works \cite{Bellaire2020, Bellaire2022, Romero2025}. The experiments were conducted with a Bruker NMR spectrometer (magnet: Ascend 400, console: Avance III HD 400, probe: PABBO 5.0~mm) with a magnetic field strength of 9.4~T (proton resonance frequency: 400.13~MHz) and a maximum gradient of 0.45~T~m\textsuperscript{-1}. The temperature control (uncertainty $\pm$ 0.1~K) was calibrated using a certified Pt-100 resistance thermometer (PTB, Braunschweig). The measurements were performed using 2.5~mm diffusion tubes (Deuteron GmbH) to minimize convection. The chemicals were used as received, at natural isotope abundance; details are provided in Table S5 of the Supporting Information.

The pulse sequence \texttt{stebpgp1s} \cite{Wu1995}, a stimulated echo sequence with bipolar gradients, was used as implemented in TopSpin 3.6.5 (Bruker). The Stejskal-Tanner Equation was used to calculate the self-diffusion coefficients $D_i$ \cite{Stejskal1965}:
\begin{equation}
\label{eqn:Stejskal}
\ln{\left( \frac{I}{I_0} \right)} = - D_{i} \gamma^2 \delta^2 \left( \Delta - \frac{\delta}{3}-\frac{\tau}{2}\right)g^2
\end{equation}

Here, $I$ is the signal intensity, $I_0$ is the intensity at the lowest gradient strength, $\gamma$ is the gyromagnetic ratio, $\delta$ is the gradient duration, $\Delta$ is the diffusion time, $\tau$ is the correction for bipolar gradients, and $g$ is the gradient strength. $D_i$ was obtained by fitting the equation to the measured $I/I_0$ ratios using a least-squares approach with the Python package \texttt{lmfit}\cite{lmfit}. Peak integrals were evaluated manually using MNova (Mestrelab). The experimental uncertainty $\sigma_{i}^{\text{exp}}$ was estimated from the root-mean-square error of the fit residuals, reported as a 95~\% confidence interval assuming a t-distribution. The uncertainty is indicated with the experimental results.

The pulse sequence parameters were $\Delta = 50$ ms and $\tau = 0.2$ ms. The gradient strengths $g$ were varied from 0.023 to 0.431~T~m\textsuperscript{-1} in eight increments with equal squared spacing. 32 scans were conducted at each increment. The gradient duration $\delta$ was adjusted (300 - \SI{5000}{\micro \second}) to ensure at least 80~\% signal attenuation from lowest to highest $g$. 

Solutions with three different solute concentrations (0.005, 0.01, and 0.025 mol mol\textsuperscript{-1}) were gravimetrically prepared for each measured solute-solvent system and measured at three temperatures (298~K, 313~K, and 333~K). When multiple peaks were present for the same compound, their respective measured $D_i$ values were averaged. The diffusion coefficients measured at the three concentrations were linearly extrapolated to infinite dilution of the solute to obtain $D^{\infty\text{,exp}}_{ij}(T)$. The overall uncertainty $\sigma^{\infty\text{,exp}}_{ij}(T)$ was calculated by combining propagated measurement errors and extrapolation uncertainty, reported as a 95~\% confidence interval assuming a t-distribution.

\section{Results and Discussion}

\subsection{Accuracy of Diffusion Coefficient Prediction}

\begin{figure}[H]
    \centering
    \includegraphics[width=0.9\textwidth]{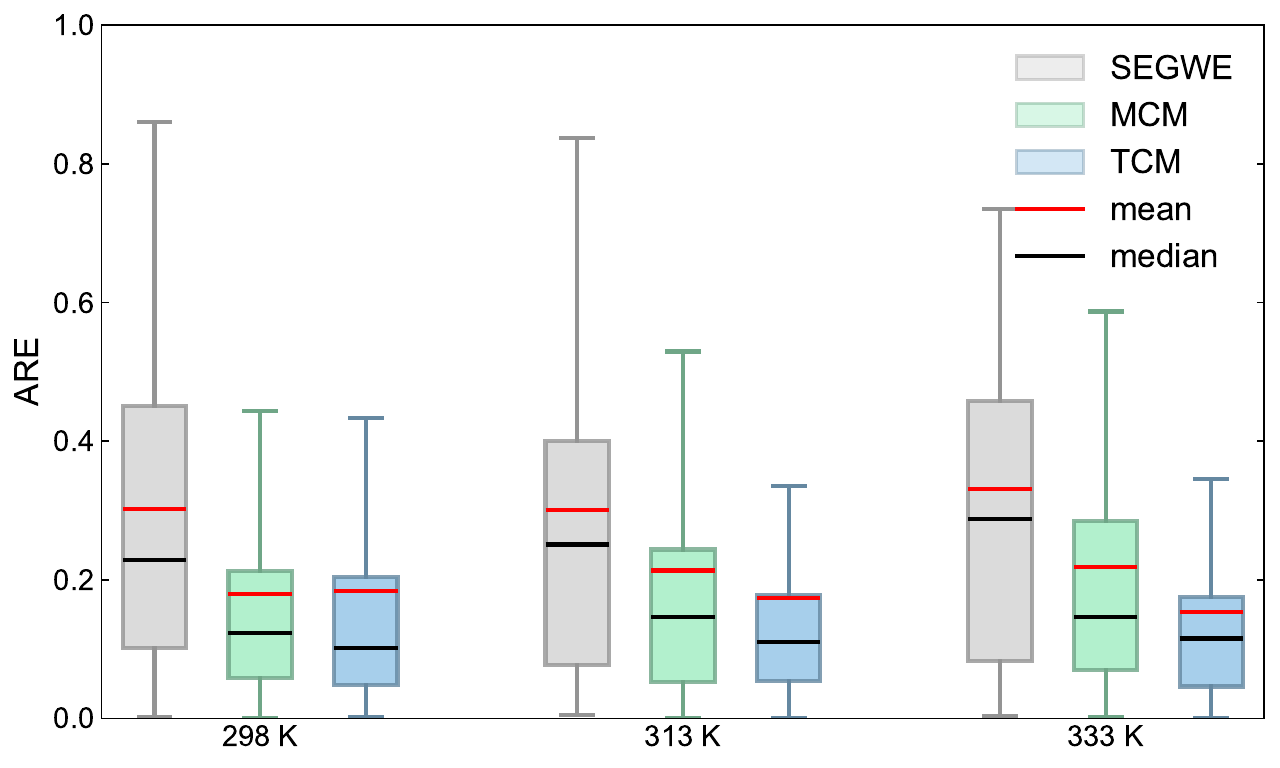}
    \caption{Boxplot of the ARE$_{ij}$ of the predicted $D_{ij}^\infty$ with SEGWE \cite{Evans2018}, MCM \cite{Grossmann2022}, and the developed TCM. MCM and TCM results were obtained using leave-one-out analysis, the SEGWE model was used as proposed by the original authors\cite{Evans2018}. Boxes represent interquartile ranges (IQR) and whiskers represent 1.5 IQR. }
    \label{fig:prediction_errors}
\end{figure}

In Figure \ref{fig:prediction_errors}, the performance of the hybrid TCM developed in this work for predicting $D_{ij}^\infty$ is compared to that of the semiempirical SEGWE \cite{Evans2018} model and that of the isothermal MCM models \cite{Grossmann2022} in terms of the ARE$_{ij}$ based on the discrete-temperature dataset compiled in this work from the literature. Note that the results of the MCMs at 313~K and 333~K shown in Figure~\ref{fig:prediction_errors} include some predictions for systems containing components, for which no experimental data were part of the training set, which is a consequence of the leave-one-out analysis and the fact that the MCMs are trained only on data for a single temperature. Consequently, the MCMs could infer the latent features of some components only from the synthetic SEGWE data. In Figure S3 in the Supporting Information, an analogous plot is shown, including only systems whose components appeared in all training data sets; the results are very similar.


Figure \ref{fig:prediction_errors} demonstrates that the MCMs (green) yield substantially lower prediction errors than SEGWE (red) \cite{Evans2018}, confirming previous results at 298~K reported by our group \cite{Grossmann2022}. Notably, the MCM approach maintains superior performance over SEGWE, also at elevated temperatures (313~K and 333~K), despite the significantly lower availability of experimental data at these temperatures.

The TCM predictions exhibit lower error scores, including narrower interquartile ranges (IQRs) and 1.5×IQR whiskers than both SEGWE and MCM, indicating that the TCM predictions are more robust and have fewer outliers than the previously available methods. This robustness in predictive performance is further illustrated in the histograms of the relative prediction errors of $D_{ij}^\infty$ for each method and temperature shown in Figure S4 in the Supporting Information.

The TCM developed in this work (blue) further improves the predictive accuracy over both SEGWE and MCM across all three studied temperatures. This result is astonishing, as one could have expected a deterioration going from an individual fit for each temperature to a global fit over all temperatures. However, the results demonstrate that incorporating diffusion coefficient data across multiple temperatures into a model's training not only broadens the model's predictive scope but also enhances its accuracy at individual temperatures.

\subsection{Analysis of the Temperature Features}

\begin{figure}[H]
    \centering
    \includegraphics[width=0.65\textwidth]{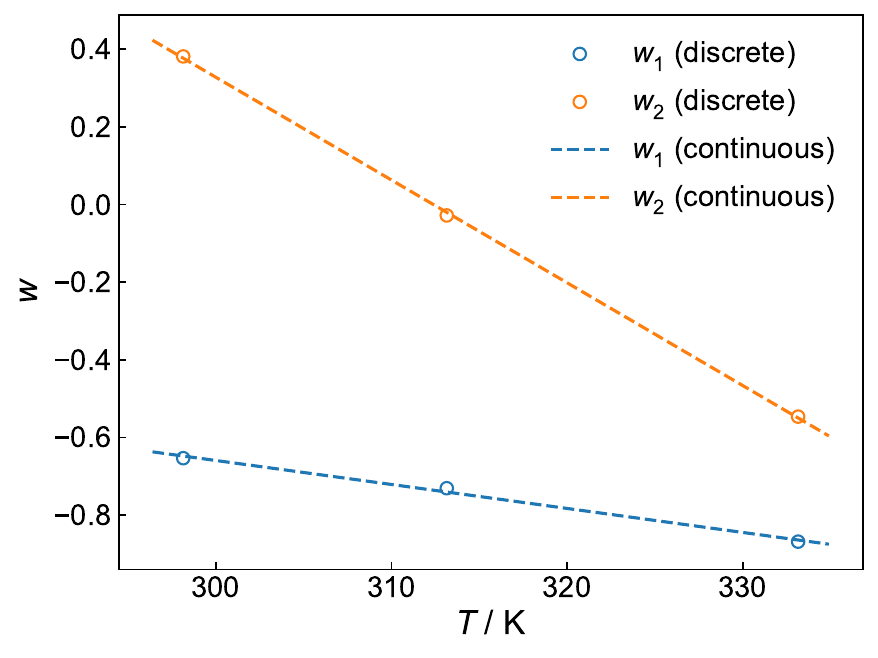}
    \caption{Temperature features $w_1(T)$ and $w_2(T)$ of a TCM trained on the full discrete data set as a function of $T$ and linear fits.}
    \label{fig:t_corr}
\end{figure}

Figure \ref{fig:t_corr} shows the latent temperature features $w_\gamma$ (or rather their means) calculated by the TCM trained on the entire discrete data set as a function of $T$.

Figure \ref{fig:t_corr} shows a linear dependence of the discrete $w_1$ and $w_2$ (circles) on the temperature $T$. With the goal of predicting $D_{ij}^\infty$ across a continuous $T$ range, we thus model the temperature dependence of $w_\gamma$ using the following equation (\ref{eqn:t_corr}):
\begin{equation}
\label{eqn:t_corr}
 w_{\gamma}(T) = A_\gamma + B_\gamma T
\end{equation}

The linear regression statistics obtained from fitting equation (\ref{eqn:t_corr}) to the discrete $w_\gamma$ are detailed in Table \ref{tab:regression}, including the coefficient of determination ($R^2$) and mean squared error (MSE) of the fit.

\begin{table}[H]
    \centering
    \begin{tabular}{lcccc}
         \toprule
         $\gamma$   & $A_\gamma$           & $B_\gamma$          & $R^2$         & MSE      \\
         \midrule
         $1$ & 1.195  & -6.179 $\cdot 10^{-3}$ & 0.9939 & 4.81 $\cdot 10^{-5}$ \\
         $2$ & 8.271  & -2.648 $\cdot 10^{-2}$ & 0.9997 & 3.18 $\cdot 10^{-5}$ \\
         \bottomrule
    \end{tabular}
    \caption{Regression statistics obtained from fitting equation (\ref{eqn:t_corr}) to the discrete $w_\gamma$, cf. Figure~\ref{fig:t_corr}.}
    \label{tab:regression}
\end{table}

Table \ref{tab:regression} shows a very strong ($R^2>0.99$) linear correlation between the $w_\gamma$ and $T$. Considering the temperature independence of the solute and solvent features $u$ and $v$ and the core tensor $\kappa$, this implies that within the considered temperature range $\ln D_{ij}^\infty$ is linear in $T$. This result does not directly correlate with Stokes-Einstein theory \cite{Sutherland1905} or SEGWE \cite{Evans2018}, as they require the solvent viscosity, the temperature dependence of which is not easily described.

\begin{figure}[H]
    \centering
    \includegraphics[width=0.9\textwidth]{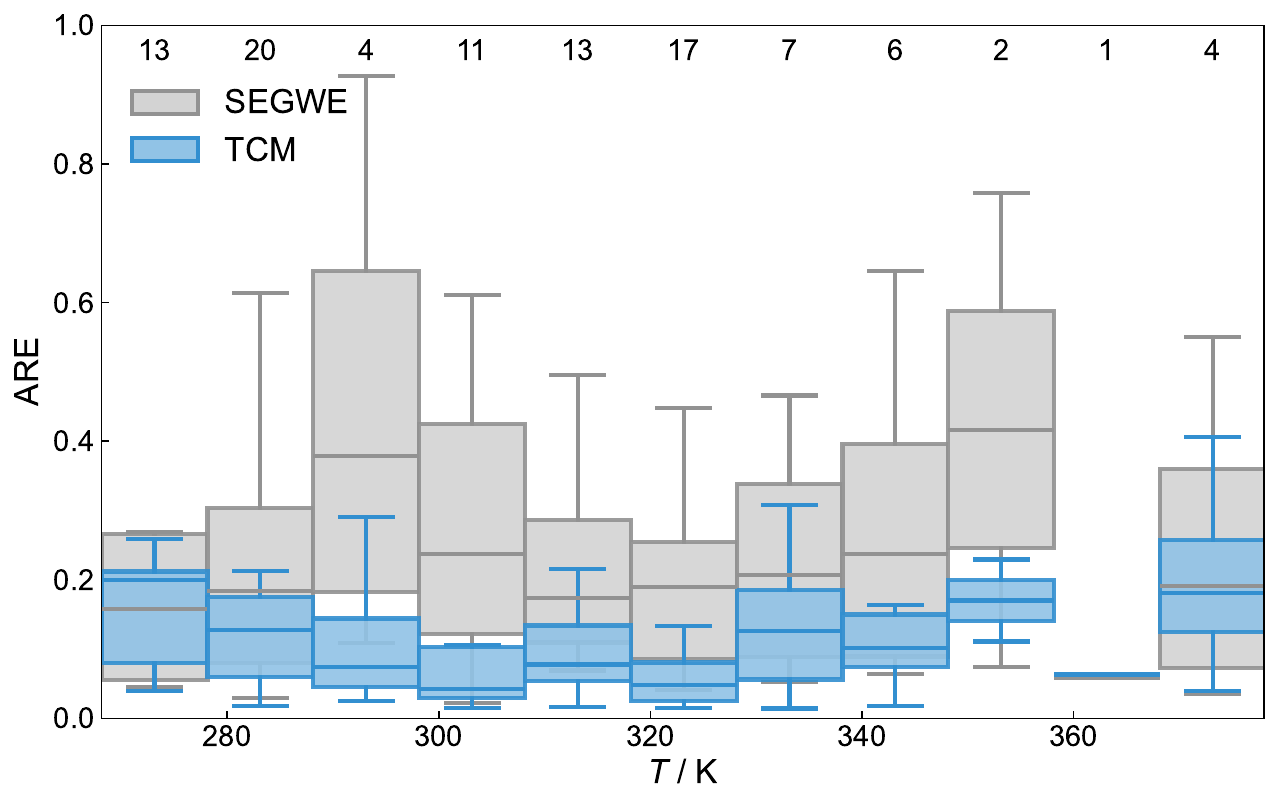}
    \caption{Boxplot of the ARE$_{ij}$ of the predicted $D_{ij}^\infty$ with SEGWE\cite{Evans2018} and the TCM as a function of $T$. The numbers above each box indicate the number of data points per bin, horizontal lines represent the median, boxes represent the IQR, and whiskers represent 1.5 IQR.}
    \label{fig:prediction_over_T}
\end{figure}

It is worth repeating that the TCM learned this correlation only from the experimental values for $D_{ij}^\infty$ at 298 K, 313 K, and 333 K. No information on the actual physical temperature was provided, as was also the case for solutes and solvents. It is thus most astonishing that the TCM was able to learn a correlation between its parameters and the temperature purely from experimental data. Using this correlation, we predicted $D_{ij}^\infty$ for the same solute-solvent matrix of Figure \ref{fig:db}, at temperatures between 268 and 378~K, the prediction error of which is shown as a function of $T$ in Figure \ref{fig:prediction_over_T}, using 10~K temperature bins.

The TCM maintains high performance across the temperature range from 268 to 378 K, with a total rMAE of 0.118, while the SEGWE model has a total rMAE of 0.263 for the same data set. Additionally, the TCM substantially outperforms SEGWE \cite{Evans2018}, even at temperatures not present in TCM's training set. It is most surprising that, despite the training data containing only a very small temperature range (35~K) and only three temperatures within it, the TCM can extrapolate easily to any unseen temperature in a much broader range (110~K), with barely any loss in accuracy. As expected with increasing temperatures, the prediction accuracy gradually worsens; thus, the linear scaling with $T$ should be used only within the specified 268~K to 378~K range. To improve prediction accuracy at higher temperatures, alternative scaling and the inclusion of experimental data at higher temperatures could be used.

\subsection{Improvement of TCM by Active Learning}
\label{ch:exp}

In Table \ref{tab:diff_inf}, the experimental diffusion coefficients at infinite dilution $D^{\infty\text{,exp}}_{ij}$ at 298~K, 313~K, and 333~K measured in this work by PFG NMR spectroscopy are reported with their respective uncertainties. The complete list of measured self-diffusion coefficients $D_i$ at the studied finite concentrations, from which the $D^{\infty\text{,exp}}_{ij}$ were derived by extrapolation, is reported in the Supporting Information. In total, $D^{\infty\text{,exp}}_{ij}$ in 19 different binary mixtures were measured.

\begin{table}[H]
\centering
\caption{Liquid-phase diffusion coefficients at infinite dilution $D^{\infty\text{,exp}}_{ij}$ measured by PFG NMR spectroscopy in this work, including experimental uncertainty $\sigma_{ij}^{\text{exp}}$. The systems are sorted according to the order in which they were selected by the AL strategy, measured, and subsequently included in the TCM training.}
\resizebox{\textwidth}{!}{%
\begin{tabular}{cllccc}
\toprule
             &                     &                  & \multicolumn{3}{c}{$D^{\infty\text{,exp}}_{ij} \: / \: 10^{-9} \text{m}^2 \text{s}^{-1} $}                               \\ \cline{4-6}
No. & Solute $i$     & Solvent $j$ & 298~K & 313~K & 333~K \\
\midrule
1  & Methyl isopropyl ketone           & 1,2-Propanediol     & 0.052 $\pm$ 0.002 & 0.115 $\pm$ 0.009 & 0.248 $\pm$ 0.044 \\
2  & Butyl acetate                     & 1,2-Propanediol     & 0.059 $\pm$ 0.002 & 0.122 $\pm$ 0.003 & 0.267 $\pm$ 0.006 \\
3  & Benzaldehyde                      & 1,2-Propanediol     & 0.059 $\pm$ 0.009 & 0.127 $\pm$ 0.001 & 0.296 $\pm$ 0.022 \\
4  & Dimethoxymethane                  & Acetone             & 4.064 $\pm$ 0.025 & 4.894 $\pm$ 0.041 & 6.161 $\pm$ 0.102 \\
5  & Water                             & Dimethoxymethane    & 5.737 $\pm$ 0.132 & 7.085 $\pm$ 0.131  & 9.270 $\pm$ 0.177 \\
6  & 2-Methyl-2,4-pentanediol          & Acetone             & 5.019 $\pm$ 0.012 & 6.323 $\pm$ 0.024 & 8.356 $\pm$ 0.030 \\
7  & m-Cresol                          & Acetonitrile        & 2.675 $\pm$ 0.016 & 3.368 $\pm$ 0.072 & 4.357 $\pm$ 0.042 \\
8  & Water                             & 2,4,6-Trioxaheptane   & 2.559 $\pm$ 0.087 & 3.249 $\pm$ 0.126 & 3.965 $\pm$ 0.049 \\
9  & Hexafluorobenzene                 & 1-Butanol            & 0.896 $\pm$ 0.005 & 1.241 $\pm$ 0.008 & 1.828 $\pm$ 0.009 \\
10 & Chlorobenzene                     & Methyl isopropyl ketone  & 2.575 $\pm$ 0.005 & 3.140 $\pm$ 0.018 & 4.381 $\pm$ 0.200 \\
11 & Benzene                           & Butyl chloride        & 3.292 $\pm$ 0.021 & 3.957 $\pm$ 0.035 & 4.994 $\pm$ 0.058 \\
12 & Methyl isopropyl ketone           & Acetone              & 3.736 $\pm$ 0.008 & 4.448 $\pm$ 0.020 & 5.495 $\pm$ 0.039 \\
13 & Glycerol                          & Acetonitrile       & 2.668 $\pm$ 0.039 & 3.219 $\pm$ 0.081  & 4.099 $\pm$ 0.042 \\
14 & Water                             & 2,4,6,8,10-Pentaoxaundecane & 0.742 $\pm$ 0.028 & 0.937 $\pm$ 0.046 & 1.276 $\pm$ 0.054 \\
15 & Water                             & 2,4,6,8-Tetraoxanonane  & 1.327 $\pm$ 0.005 & 1.730 $\pm$ 0.034 & 2.351 $\pm$ 0.024 \\
16 & 2,4,6-Trioxaheptane & Dimethoxymethane  & 3.464 $\pm$ 0.041 & 4.186 $\pm$ 0.026 & 4.869 $\pm$ 0.026 \\
17 & Butyric acid                      & Dimethoxymethane      & 2.381 $\pm$ 0.109 & 3.262 $\pm$ 0.015 & 4.175 $\pm$ 0.026 \\
18 & Di-tert-butylsulfide              & Dimethoxymethane     & 2.757 $\pm$ 0.008 & 3.347 $\pm$ 0.010 & 4.306 $\pm$ 0.025 \\
19 & Phenol                            & Dimethoxymethane    & 2.768 $\pm$ 0.012 & 3.418 $\pm$ 0.018 & 4.470 $\pm$ 0.045 \\
\bottomrule
\end{tabular}
}
\label{tab:diff_inf}
\end{table}

As expected, $D^{\infty\text{,exp}}_{ij}$ increases with increasing temperature for all systems studied. The experimental uncertainty shows significant variation. In some cases, the uncertainty is as high as 10 \% (partly even higher), mainly caused by the decreasing sensitivity of the NMR experiment at high temperatures in combination with low solute concentrations.

In the experimental implementation of the AL approach, we observed that some solvents were proposed particularly often by uncertainty sampling. In this study, as shown in Table \ref{tab:diff_inf}, 1,2-propanediol was initially suggested most often, which we thus excluded from the sampling pool after the third measurement. The same effect was observed for dimethoxymethane towards the end of our measurements.

Using the entire set of new data for 19 previously unstudied mixtures for the training, the prediction rMSE decreased from 0.18 to 0.15 at 298~K, from 0.10 to 0.08 at 313~K, and from 0.07 to 0.06 at 333~K, while the occupation rate of the matrix increased only by 1.8 \%. The prediction rMAE and rMSE after each AL iteration is shown in the Supporting Information Figure S5.  These results show that substantial improvements in the prediction of diffusion coefficients can be achieved with only a few additional measurements, consistent with our previous findings~\cite{Romero2025}. 

In the Supporting Information, we report a set of parameters obtained with the TCM after training on all literature data (cf. Figure \ref{fig:db}) and the new data measured in this work.


\section{Conclusions}

In this work, we have introduced a novel tensor completion method (TCM) for predicting the diffusion coefficients at infinite dilution, $D_{ij}^\infty$, in binary systems at different temperatures. The method is trained on $D_{ij}^\infty$ data at 298~K, 313~K, and 333~K, but allows predictions at any temperature between 268~K and 378~K. Thereby, we extended previously available matrix completion methods (MCMs), which were restricted to the isothermal case. The TCM achieves significantly higher prediction accuracies than the semi-empirical SEGWE model \cite{Evans2018}. The global TCM model also provides better predictions than MCMs trained individually at different temperatures, indicating that the TCM benefits from joint training across different temperatures while simultaneously enabling generalization over a large temperature range.

Furthermore, the available data on $D_{ij}^\infty$ were extended through measurements using PFG NMR spectroscopy, in which the systems were selected using an active learning (AL) approach guided by the model's uncertainty. In total, 19 systems for which no prior data were available were measured at 298~K, 313~K, and 333~K. Even though this only increases the tensor's occupation rate by 1.8 \%, considerable improvements in prediction quality were observed. However, further improvements could be achieved by developing tailored query strategies in future work.

\section{Data Availability Statement}

Most of the experimental data on $D_{ij}^\infty$ used for training and testing in this work were used under license for this study; they are available directly from Dortmund Data Bank (DDB)\cite{ddb} version 2025. Additional experimental data on $D_{ij}^\infty$ found during our comprehensive literature studied are available from their original sources\cite{ddb,tominaga1984limiting,tominaga1990diffusion,Tominaga1990,Hsieh2012,Li1990,Schatzberg1965,Lin2008, hurle1982tracer,tyn1975temperature,sanni1971diffusion,bonoli1968diffusion,te1995diffusion,alizadeh1982mutual,anderson1958mutual,hashim2007diffusion,yumet1985tracer,chen1983tracer,clark1986mutual,Mross2024,Wagner2025,Romero2025}. In the Supporting Information of this work, we report the new diffusion data measured in this work, the complete Stan code used in processing the data sets in this work, as well as a set of parameters obtained with the TCM after training on all literature data (cf. Figure \ref{fig:db}) and the new data measured in this work.

\section{Conflicts of Interest}

There are no conflicts of interest to declare.

\begin{acknowledgement}

We gratefully acknowledge financial support by the Carl Zeiss Foundation in the frame of the project "Process Engineering 4.0" and by DFG in the frame of the Research Training Group GRK 2908 "Valuable Wastewater (WERA)" (grant number 503479768) and the Priority Program SPP 2363 "Molecular Machine Learning" (grant number 497201843). Furthermore, FJ gratefully acknowledges financial support by DFG in the frame of the Emmy-Noether program (grant number 528649696).

\end{acknowledgement}






\bibliography{achemso-demo}

\end{document}


\section{Diffusion Coefficient Data Availability}

\begin{figure}[H]
\centering
\includegraphics[width=1\textwidth]{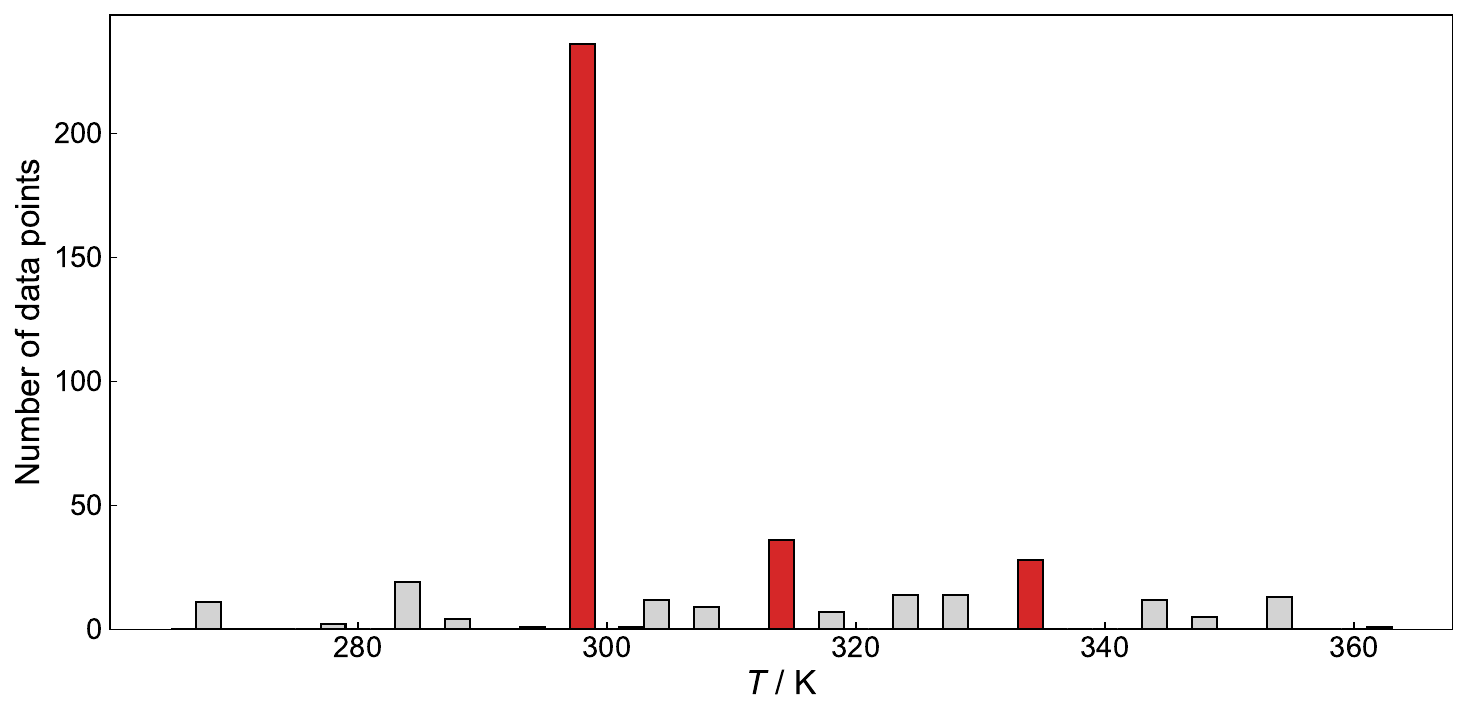}
\caption{Availability of experimental data for liquid-phase diffusion coefficients $D_{ij}^{\infty\text{,exp}}$ of solutes $i$ in solvents $j$ at infinite dilution at temperatures 298, 313, and 333~K from the DDB\cite{ddb} and our previous work\cite{Romero2025,Mross2024,Bellaire2022}.}
\label{fig:diff_of_T}
\end{figure}

In this work, we have extended the database of Großmann et al.\cite{Grossmann2022} with new $D_{ij}^\infty$ from the 2025 version of the DDB \cite{ddb} and data from our previous works~\cite{Romero2025,Mross2024,Bellaire2022}. The number of available data points per 2~K temperature bin is shown as a function of temperature $T$ in Figure \ref{fig:diff_of_T}. The three temperatures for which the most data are available are 298~K, 313~K, and 333~K, which are marked in red in Figure~\ref{fig:diff_of_T}.

\section{Compound Identifiers}
\label{ident}
\setcounter{page}{1}

Tables \ref{tab:Solvents} to \ref{tab:solutes_al} list the identifiers for solvents and solutes, respectively, used in Figures \ref{fig:db} and \ref{fig:db_al}. Compounds are also listed using simplified molecular input line entry system (SMILES) notation generated using RDKit \cite{rdkit}.

\begin{center}
\begin{spacing}{1.0}
\begin{longtable}{lll}
\caption{Identifiers for solutes considered in the full literature database, cf. Figure \ref{fig:db}, including their SMILES.} \\
\hline Nr. & Solute                      & SMILES                 \\ \hline
\endfirsthead

\multicolumn{3}{c}%
{{\tablename\ \thetable{} -- continued from previous page}} \\
\hline Nr. & Solute                      & SMILES                                   \\ \hline 
\endhead

\hline
\endfoot

\hline
\endlastfoot

1  & Methane                             & C                      \\
2  & Water                               & O                      \\
3  & Methanol                            & CO                     \\
4  & Argon                               & {[}Ar{]}               \\
5  & Acetonitrile                        & CC\#N                  \\
6  & Carbon dioxide                      & O=C=O                  \\
7  & Ethanol                             & CCO                    \\
8  & Acetone                             & CC(C)=O                \\
9  & Acetic acid                         & CC(=O)O                \\
10 & 1-Propanol                          & CCCO                   \\
11 & Propionic acid                      & CCC(=O)O               \\
12 & 1-Butanol                           & CCCCO                  \\
13 & Dimethoxymethane                    & COCOC                  \\
14 & Benzene                             & c1ccccc1               \\
15 & Krypton                             & {[}Kr{]}               \\
16 & Cyclohexane                         & C1CCCCC1               \\
17 & Methyl isopropyl ketone             & CC(=O)C(C)C            \\
18 & Hexane                              & CCCCCC                 \\
19 & Ethyl acetate                       & CCOC(C)=O              \\
20 & Butyric acid                        & CCCC(=O)O              \\
21 & Butyl chloride                      & CCCCCl                 \\
22 & Glycerol                            & OCC(O)CO               \\
23 & Toluene                             & Cc1ccccc1              \\
24 & Phenol                              & Oc1ccccc1              \\
25 & Heptane                             & CCCCCCC                \\
26 & Benzaldehyde                        & O=Cc1ccccc1            \\
27 & 2,4,6-Trioxaheptan                  & COCOCOC                \\
28 & Ethylbenzene                        & CCc1ccccc1             \\
29 & Benzyl alcohol                      & OCc1ccccc1             \\
30 & m-Cresol                            & Cc1cccc(O)c1           \\
31 & Chlorobenzene                       & Clc1ccccc1             \\
32 & Butyl acetate                       & CCCCOC(C)=O            \\
33 & 2-Methyl-2,4-pentanediol            & C{[}C@H{]}(O)CC(C)(C)O \\
34 & Acetophenone                        & CC(=O)c1ccccc1         \\
35 & p-Chlorotoluene                     & Cc1ccc(Cl)cc1          \\
36 & Naphthalene                         & c1ccc2ccccc2c1         \\
37 & Xenon                               & {[}Xe{]}               \\
38 & 2,4,6,8-Tetraoxanonane              & COCOCOCOC              \\
39 & Methyl iodide                       & CI                     \\
40 & Tetrachloromethane                  & ClC(Cl)(Cl)Cl          \\
41 & 2,4,6,8,10-Pentaoxaundecane         & COCOCOCOCOC            \\
42 & Dodecane                            & CCCCCCCCCCCC           \\
43 & Di-tert-butylsulfide                & CC(C)(C)SSC(C)(C)C     \\
44 & Hexafluorobenzene                   & Fc1c(F)c(F)c(F)c(F)c1F \\
45 & Hexadecane                          & CCCCCCCCCCCCCCCC                      

\label{tab:Solutes}
\end{longtable}
\end{spacing}
\end{center}

\begin{table}[H]
\caption{Identifiers for solvents considered in the full literature database, cf. Figure \ref{fig:db}, including their SMILES.}
\centering
\label{tab:Solvents}
\begin{tabular}{lll}

\toprule
Nr. & Solvent                     & SMILES                 \\
\midrule
1  & Hexane                              & CCCCCC                 \\
2  & Acetone                             & CC(C)=O                \\
3  & Dimethoxymethane                    & COCOC                  \\
4  & Acetonitrile                        & CC\#N                  \\
5  & Heptane                             & CCCCCCC                \\
6  & Butyl chloride                      & CCCCCl                 \\
7  & Ethyl acetate                       & CCOC(C)=O              \\
8  & Methyl isopropyl ketone             & CC(=O)C(C)C            \\
9  & Methanol                            & CO                     \\
10 & Chloroform                          & ClC(Cl)Cl              \\
11 & Toluene                             & Cc1ccccc1              \\
12 & 2,4,6-Trioxaheptane                 & COCOCOC                \\
13 & Benzene                             & c1ccccc1               \\
14 & Butyl acetate                       & CCCCOC(C)=O            \\
15 & Decane                              & CCCCCCCCCC             \\
16 & Hexafluorobenzene                   & Fc1c(F)c(F)c(F)c(F)c1F \\
17 & Cyclohexane                         & C1CCCCC1               \\
18 & Tetrachloromethane                  & ClC(Cl)(Cl)Cl          \\
19 & Water                               & O                      \\
20 & 2,4,6,8-Tetraoxanonane              & COCOCOCOC              \\
21 & Ethanol                             & CCO                    \\
22 & Dodecane                            & CCCCCCCCCCCC           \\
23 & 2,4,6,8,10-Pentaoxaundecane         & COCOCOCOCOC            \\
24 & N-Methyl-2-pyrrolidone              & CN1CCCC1=O             \\
25 & 1-Propanol                          & CCCO                   \\
26 & 2-Propanol                          & CC(C)O                 \\
27 & Tetradecane                         & CCCCCCCCCCCCCC         \\
28 & 1-Butanol                           & CCCCO                  \\
29 & Hexadecane                          & CCCCCCCCCCCCCCCC       \\
30 & 1-Octanol                           & CCCCCCCCO              \\
31 & 1,2-Propanediol                     & C{[}C@@H{]}(O)CO      \\
\bottomrule
\end{tabular}
\end{table}

\begin{center}
\begin{spacing}{1.0}
\begin{longtable}{lll}
\caption{Identifiers for and solutes considered in the reduced database used for the active learning (AL) portion, cf. Figure \ref{fig:db_al} including SMILES.} \\
\hline Nr. & Solute                      & SMILES                 \\ \hline
\endfirsthead

\multicolumn{3}{c}%
{{\tablename\ \thetable{} -- continued from previous page}} \\
\hline Nr. & Solute                      & SMILES                                   \\ \hline 
\endhead

\hline
\endfoot

\hline
\endlastfoot

1   & Water                       & O                      \\
2   & Methanol                    & CO                     \\
3   & Acetonitrile                & CC\#N                  \\
4   & Ethanol                     & CCO                    \\
5   & Acetone                     & CC(C)=O                \\
6   & Acetic acid                 & CC(=O)O                \\
7   & 1-Propanol                  & CCCO                   \\
8   & Propionic acid              & CCC(=O)O               \\
9   & 1-Butanol                   & CCCCO                  \\
10  & Dimethoxymethane            & COCOC                  \\
11  & Benzene                     & c1ccccc1               \\
12  & Cyclohexane                 & C1CCCCC1               \\
13  & Methyl isopropyl ketone     & CC(=O)C(C)C            \\
14  & Hexane                      & CCCCCC                 \\
15  & Butyric acid                & CCCC(=O)O              \\
16  & Ethyl acetate               & CCOC(C)=O              \\
17  & Glycerol                    & OCC(O)CO               \\
18  & Toluene                     & Cc1ccccc1              \\
19  & Butyl chloride              & CCCCCl                 \\
20  & Phenol                      & Oc1ccccc1              \\
21  & Heptane                     & CCCCCCC                \\
22  & 2,4,6-Trioxaheptane         & COCOCOC                \\
23  & Benzaldehyde                & O=Cc1ccccc1            \\
24  & Ethylbenzene                & CCc1ccccc1             \\
25  & 3-Methylphenol              & Cc1cccc(O)c1           \\
26  & Benzyl alcohol              & OCc1ccccc1             \\
27  & Chlorobenzene               & Clc1ccccc1             \\
28  & Butyl acetate               & CCCCOC(C)=O            \\
29  & 2-Methyl-2,4-pentanediol    & C{[}C@H{]}(O)CC(C)(C)O \\
30  & Acetophenone                & CC(=O)c1ccccc1         \\
31  & p-Chlorotoluene             & Cc1ccc(Cl)cc1          \\
32  & Naphthalene                 & c1ccc2ccccc2c1         \\
33  & 2,4,6,8-Tetraoxanonane       & COCOCOCOC              \\
34  & Iodomethane                 & CI                     \\
35  & Di-\textit{tert}-butyl sulfide       & CC(C)(C)SSC(C)(C)C     \\
36  & Tetrachloromethane          & ClC(Cl)(Cl)Cl          \\
37  & 2,4,6,8,10-Pentaoxaundecane & COCOCOCOCOC            \\
38  & Dodecane                    & CCCCCCCCCCCC           \\
39  & Hexafluorobenzene           & Fc1c(F)c(F)c(F)c(F)c1F \\
40  & Hexadecane                  & CCCCCCCCCCCCCCCC                         

\label{tab:solutes_al}
\end{longtable}
\end{spacing}
\end{center}

\begin{table}[H]
\caption{Identifiers for solvents considered in the reduced database used for the AL portion, cf. Figure \ref{fig:db_al}, including SMILES.}
\centering
\label{tab:solvents_al}
\begin{tabular}{lll}

\toprule
Nr. & Solvent                     & SMILES                 \\
\midrule
1   & Hexane                      & CCCCCC                 \\
2   & Acetone                     & CC(C)=O                \\
3   & Dimethoxymethane            & COCOC                  \\
4   & Acetonitrile                & CC\#N                  \\
5   & Heptane                     & CCCCCCC                \\
6   & Butyl chloride              & CCCCCl                 \\
7   & Ethyl acetate               & CCOC(C)=O              \\
8   & Methyl isopropyl ketone     & CC(=O)C(C)C            \\
9   & Methanol                    & CO                     \\
10  & Chloroform                  & ClC(Cl)Cl              \\
11  & Toluene                     & Cc1ccccc1              \\
12  & Benzene                     & c1ccccc1               \\
13  & 2,4,6-Trioxaheptane         & COCOCOC                \\
14  & Butyl acetate               & CCCCOC(C)=O            \\
15  & Hexafluorobenzene           & Fc1c(F)c(F)c(F)c(F)c1F \\
16  & Cyclohexane                 & C1CCCCC1               \\
17  & Tetrachloromethane          & ClC(Cl)(Cl)Cl          \\
18  & Water                       & O                      \\
19  & 2,4,6,8-Tetraoxanonane      & COCOCOCOC              \\
20  & Ethanol                     & CCO                    \\
21  & Dodecane                    & CCCCCCCCCCCC           \\
22  & 2,4,6,8,10-Pentaoxaundecane & COCOCOCOCOC            \\
23  & N-Methyl-2-pyrrolidone      & CN1CCCC1=O             \\
24  & 1-Propanol                  & CCCO                   \\
25  & 1-Butanol                   & CCCCO                  \\
26  & Hexadecane                  & CCCCCCCCCCCCCCCC       \\
27  & 1,2-Propanediol             & C{[}C@@H{]}(O)CO       \\
\bottomrule
\end{tabular}
\end{table}

\section{List of Chemicals}

Table \ref{tab:chem} lists the suppliers and purities of the chemicals used in the course of this work. Deionized and purified water was prepared using an ultrapure water system (Omnia series, stakpure).

\begin{table}[H]
\centering
\caption{Suppliers and purities of the chemicals used in this work. The purities were taken from the supplier's data sheets.} \label{tab:chem}
    \begin{tabular}{lllc}
        \toprule
        Chemical & Formula & Supplier & Purity \\
        &  &  & \% \\
        \midrule
        1,2-Propanediol & C$_3$H$_8$O$_2$  & Sigma-Aldrich & $\ge$ 99.50 \\
        2-Methyl-2,4-pentanediol & C$_6$H$_{14}$O$_2$  & TCI & $>$ 99.00 \\
        Acetone & C$_3$H$_6$O & Sigma-Aldrich & $\ge$ 99.90 \\
        Acetonitrile & C$_2$H$_3$N & Sigma-Aldrich & $\ge$ 99.90 \\
        Benzaldehyde & C$_7$H$_6$O & Roth & $\ge$ 99.50 \\
        Benzene & C$_6$H$_6$ & Applichem & $\ge$ 99.50 \\
        1-Butanol & C$_4$H$_10$O & Sigma-Aldrich & $\ge$ 99.90 \\
        Butyl acetate & C$_6$H$_{12}$O$_2$ & Sigma-Aldrich & $\ge$ 99.50 \\
        Butyl chloride & C$_4$H$_9$Cl & Sigma-Aldrich & $\ge$ 99.50 \\
        Butyric acid & C$_4$H$_8$O$_2$ & TCI & $>$ 99.00 \\
        Chlorobenzene & C$_6$H$_5$Cl & Thermo Scientific & $\ge$ 99.50 \\
        Chlorotoluene & C$_7$H$_7$Cl & Acros Organics &  98.00 \\
        Dimethoxymethane & C$_3$H$_8$O & Sigma-Aldrich & 99.00 \\
        Di-\textit{tert}-butyl sulfide & C$_8$H$_{18}$S & TCI &  $>$ 99.00 \\
        Glycerol & C$_3$H$_8$O$_3$ & Sigma-Aldrich & $>$ 99.00 \\
        Hexafluorobenzene & C$_6$F$_6$ & Sigma-Aldrich & $\ge$ 99.00 \\
        $m$-Cresol & C$_7$H$_8$O & Sigma-Aldrich & 99.00 \\
        Methyl isopropyl ketone & C$_5$H$_10$O & TCI & $>$ 98.00 \\
        2,4,6,8,10-Pentaoxaundecane & C$_6$H$_14$O$_5$ & BASF SE & 98.50 \\
        Phenol & C$_6$H$_5$OH & Sigma-Aldrich & $\ge$ 99.50 \\
        2,4,6,8-Tetraoxanonane & C$_5$H$_12$O$_4$ & BASF SE & 98.50 \\
        2,4,6-Trioxaheptane & C$_4$H$_10$O$_3$ & BASF SE & 98.50 \\
        \bottomrule
    \end{tabular}
\end{table}

\section{Hyperparameter Optimization}

\begin{figure}[H]
\centering
    \begin{subfigure}{.32\textwidth}
    \centering
    \includegraphics[width=1\textwidth]{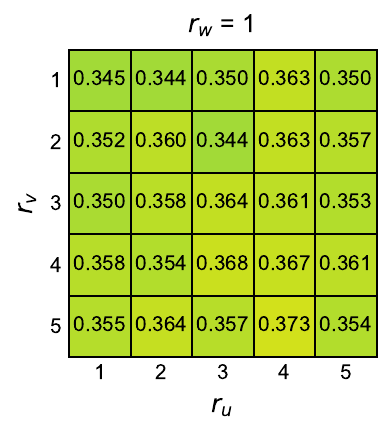}
    \end{subfigure}
    \begin{subfigure}{.32\textwidth}
    \centering
    \includegraphics[width=1\textwidth]{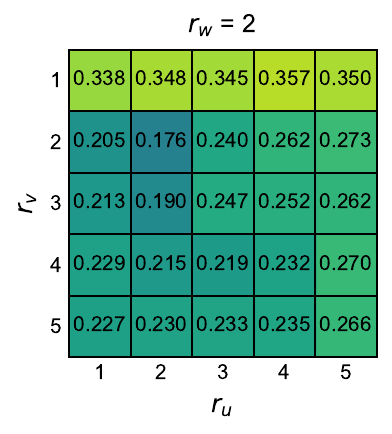}
    \end{subfigure}
    \begin{subfigure}{.32\textwidth}
    \centering
    \includegraphics[width=1\textwidth]{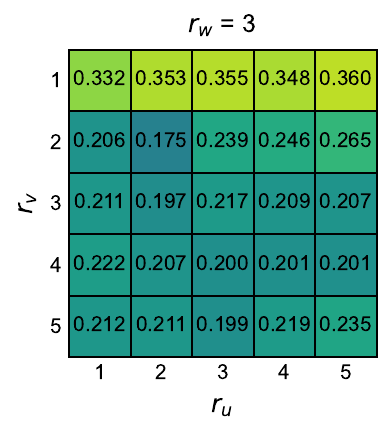}
    \end{subfigure}
    \begin{subfigure}{.32\textwidth}
    \centering
    \includegraphics[width=1\textwidth]{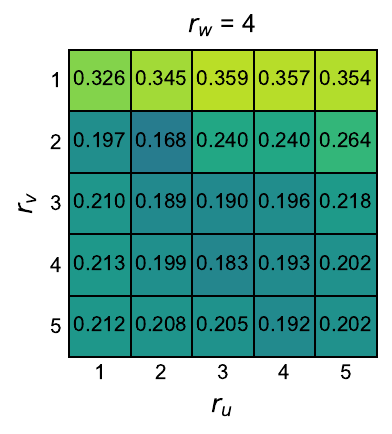}
    \end{subfigure}
    \begin{subfigure}{.32\textwidth}
    \centering
    \includegraphics[width=1\textwidth]{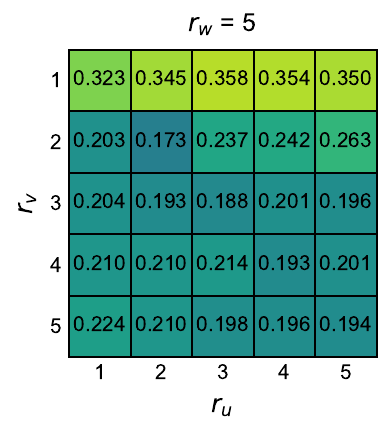}
    \end{subfigure}
    \caption{rMAE for predicting $D_{ij}^\infty$ (averaged over solutes, solvents, and temperatures) by TCM for different combinations of hyperparameters $r_u$, $r_v$, and $r_w$ evaluated by leave-one-out analysis based on the experimental data set in Figure \ref{fig:db}.}
    \label{fig:hyperparam}
\end{figure}

Figure \ref{fig:hyperparam} shows the prediction error in rMAE of the TCM for various combinations of the hyperparameters $r_u$, $r_v$, and $r_w$ evaluated by leave-one-out analysis.

Figure \ref{fig:hyperparam} shows that the lowest prediction error (rMAE = 0.168) was attained for $r_u=r_v=2$ and $r_w=4$. However, since the rMAE barely changed as a function of $r_w$ for $r_w>1$, we have chosen $r_u=r_v=r_w=2$ to keep the number of model parameters at a minimum.

\section{Diffusion Coefficient Prediction Accuracies: Boxplot}

\begin{figure}[H]
    \centering
    \includegraphics[width=0.9\textwidth]{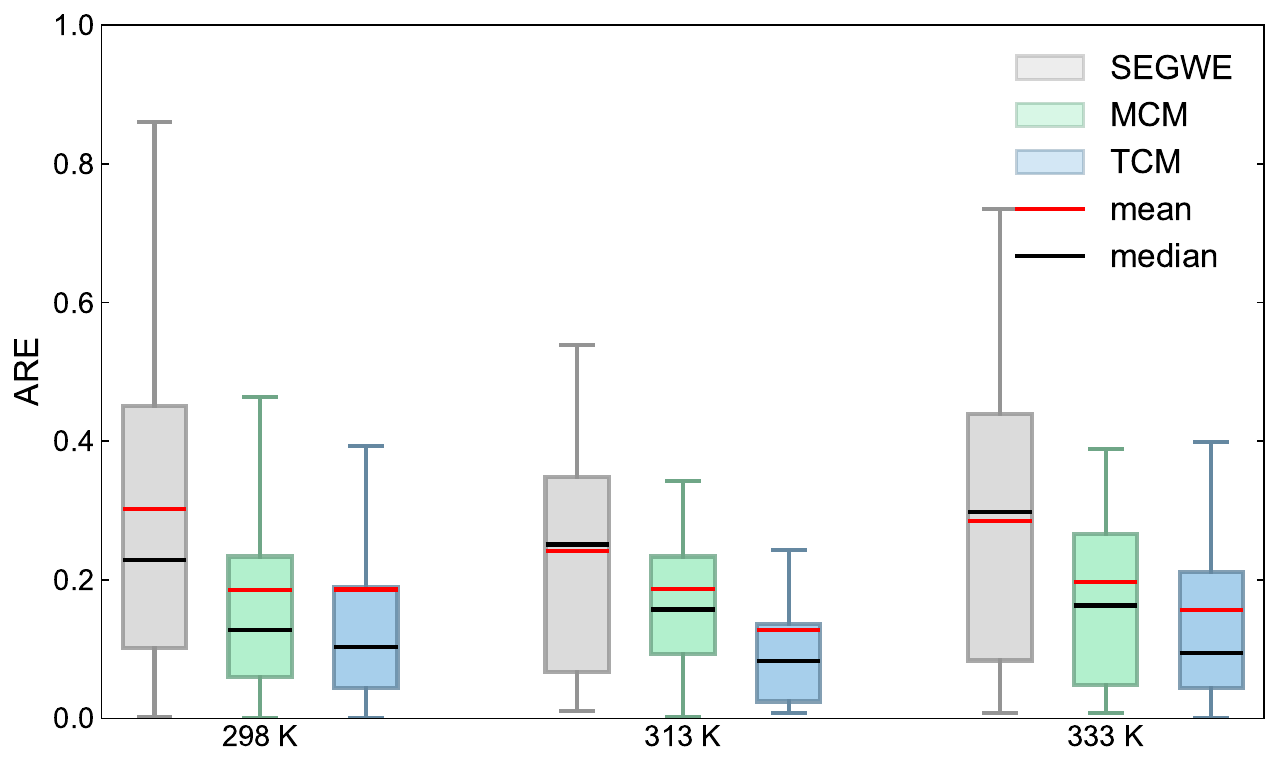}
    \caption{Boxplot of the ARE of the predicted $D_{ij}^\infty$ with SEGWE \cite{Evans2018}, MCM \cite{Grossmann2022}, and the developed TCM obtained using leave-one-out analysis. Boxes represent interquartile ranges (IQR) and whiskers represent 1.5 IQR. Evaluated based only on data points whose components appeared in all training data sets.}    \label{fig:prediction_errors_SI}
\end{figure}

In Figure \ref{fig:prediction_errors_SI}, the performance of the hybrid TCM developed in this work for predicting $D_{ij}^\infty$ is compared to that of the semi-empirical SEGWE \cite{Evans2018} model and that of the isothermal MCM models \cite{Grossmann2022} in terms of the absolute relative error (ARE) based on the discrete-temperature dataset from the literature. Compared to Figure \ref{fig:prediction_errors}, only systems whose components appeared in all training sets were included, i.e., solutes and solvents which were not available at one or two temperatures were removed. The results are very similar to the ones shown in Figure \ref{fig:prediction_errors}.

\section{Histogram Representations of Prediction Accuracy}

\begin{figure}[H]
\centering
    \begin{subfigure}{.49\textwidth}
    \centering
    \includegraphics[width=1\textwidth]{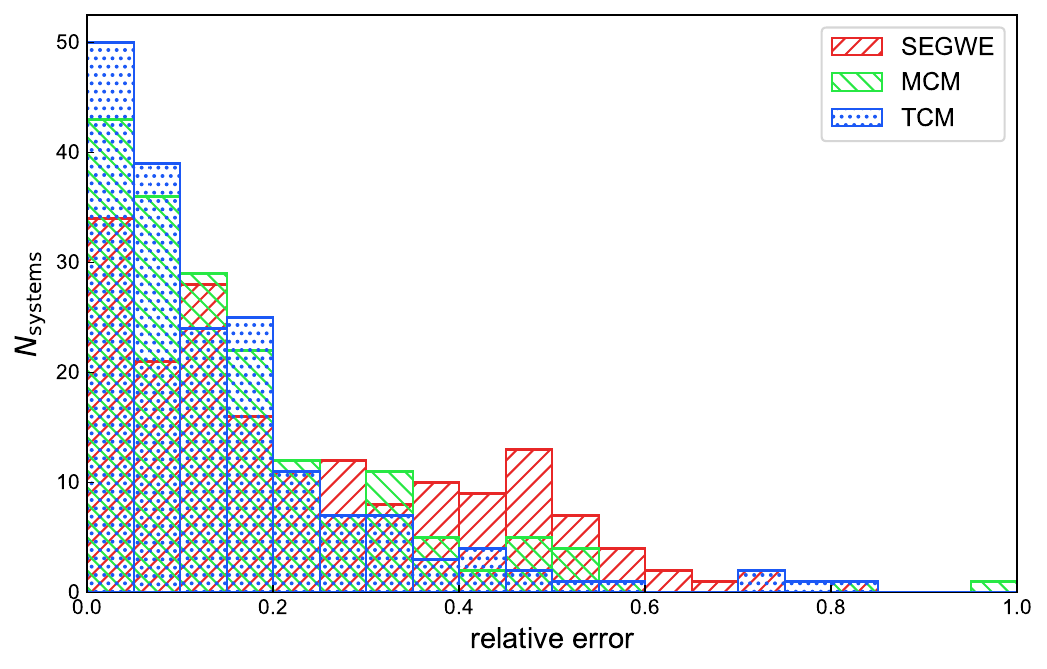}
    \caption{298~K}
    \label{fig:histo_298}
    \end{subfigure}
    \begin{subfigure}{.49\textwidth}
    \centering
    \includegraphics[width=1\textwidth]{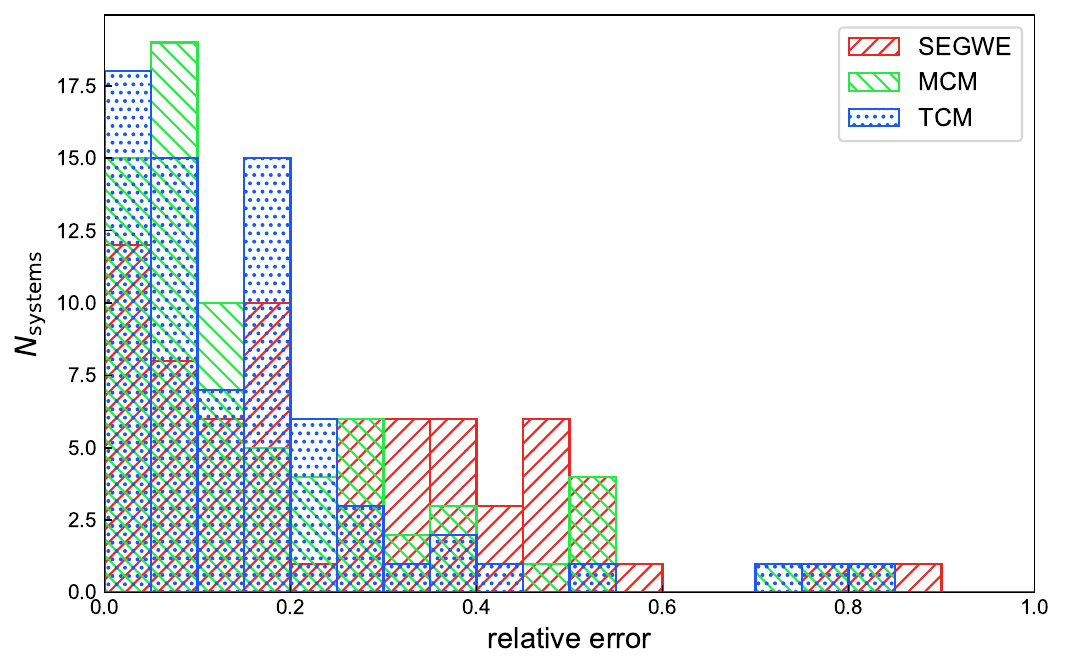}
    \caption{313~K}
    \label{fig:histo_313}
    \end{subfigure}
    \begin{subfigure}{.49\textwidth}
    \centering
    \includegraphics[width=1\textwidth]{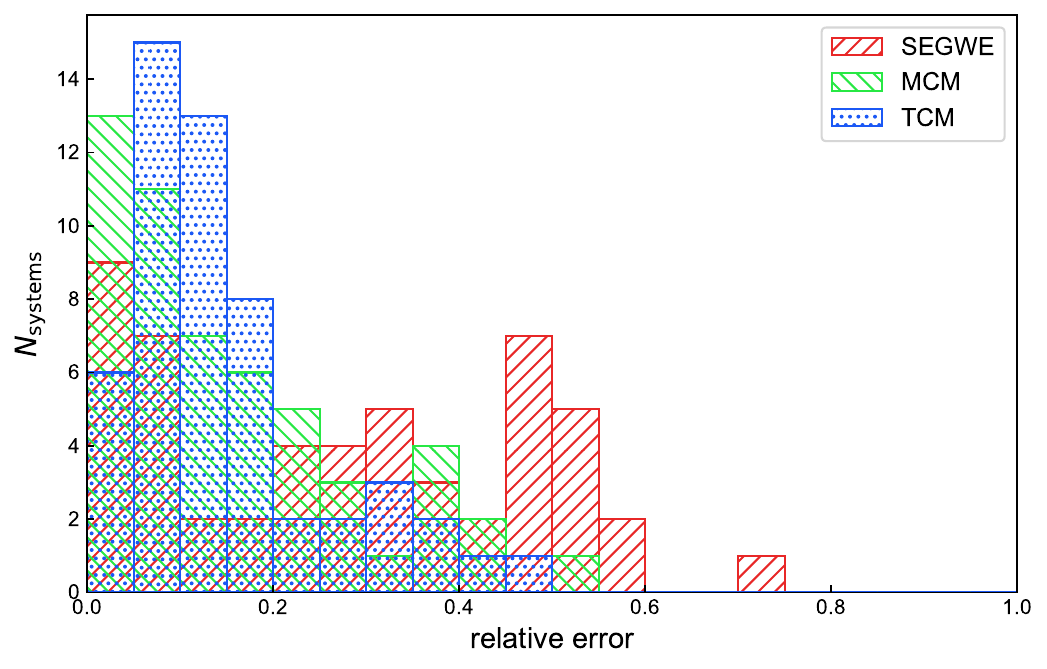}
    \caption{333~K}
    \label{fig:histo_333}
    \end{subfigure}
    \caption{Histograms showing the number of systems $N_\mathrm{systems}$ for which $D_{ij}^\infty$ is predicted with a certain relative error with SEGWE, the temperature-specific MCMs, and the developed TCM for the three studied temperatures. The results of the MCMs and the TCM were obtained using a leave-one-out analysis.}
    \label{fig:histo}
\end{figure}

Figure \ref{fig:histo} further confirms our conclusions obtained from Figure \ref{fig:prediction_errors}, showing that the TCM outperforms the prediction accuracy of the MCMs and the SEGWE model across all temperatures.




\section{Improvement of TCM by Active Learning}

\begin{figure}[H]
\centering
    \begin{subfigure}{.49\textwidth}
    \centering
    \caption{298~K}
    \includegraphics[width=1\textwidth]{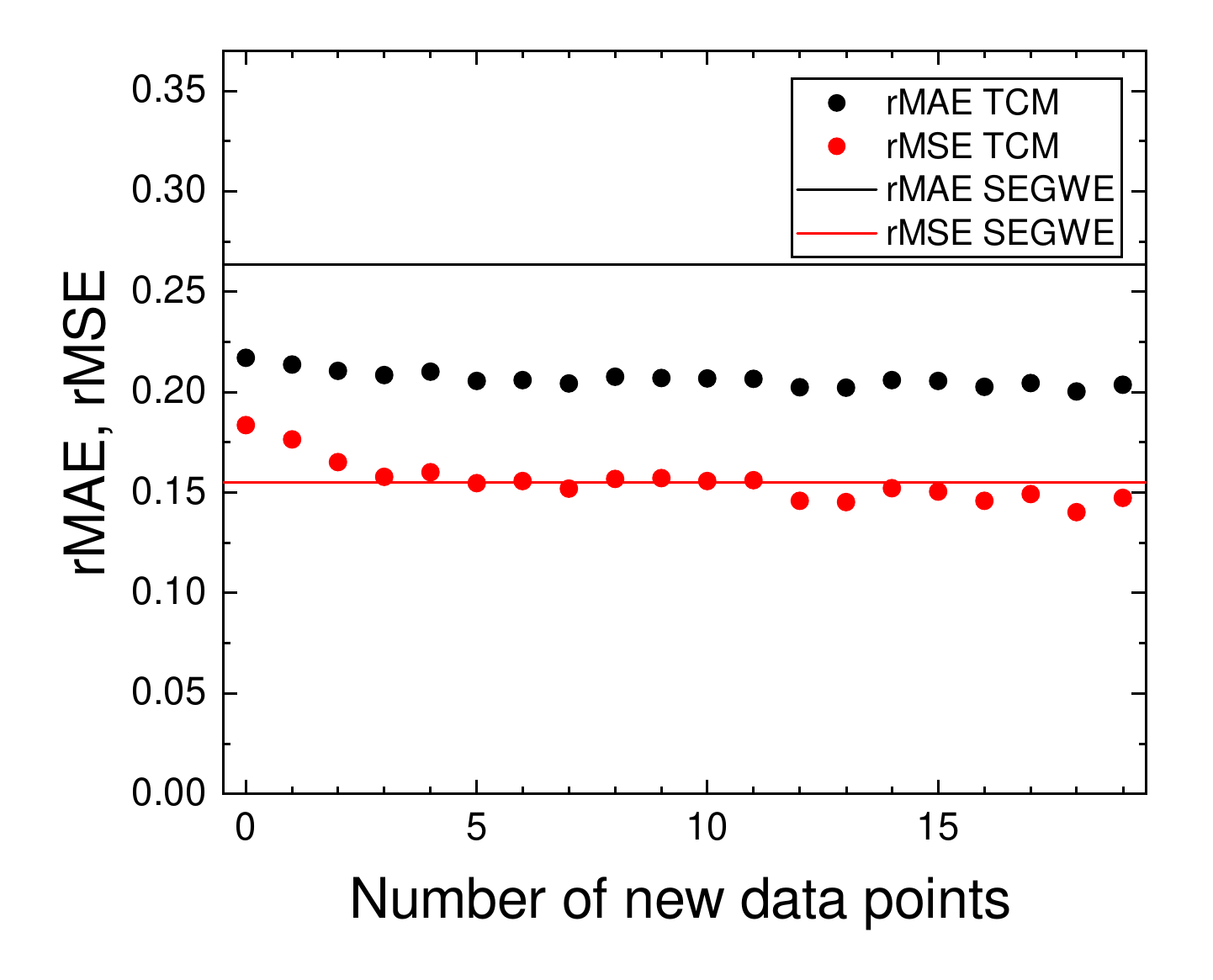}
    \end{subfigure}
    \begin{subfigure}{.49\textwidth}
    \centering
    \caption{313~K}
    \includegraphics[width=1\textwidth]{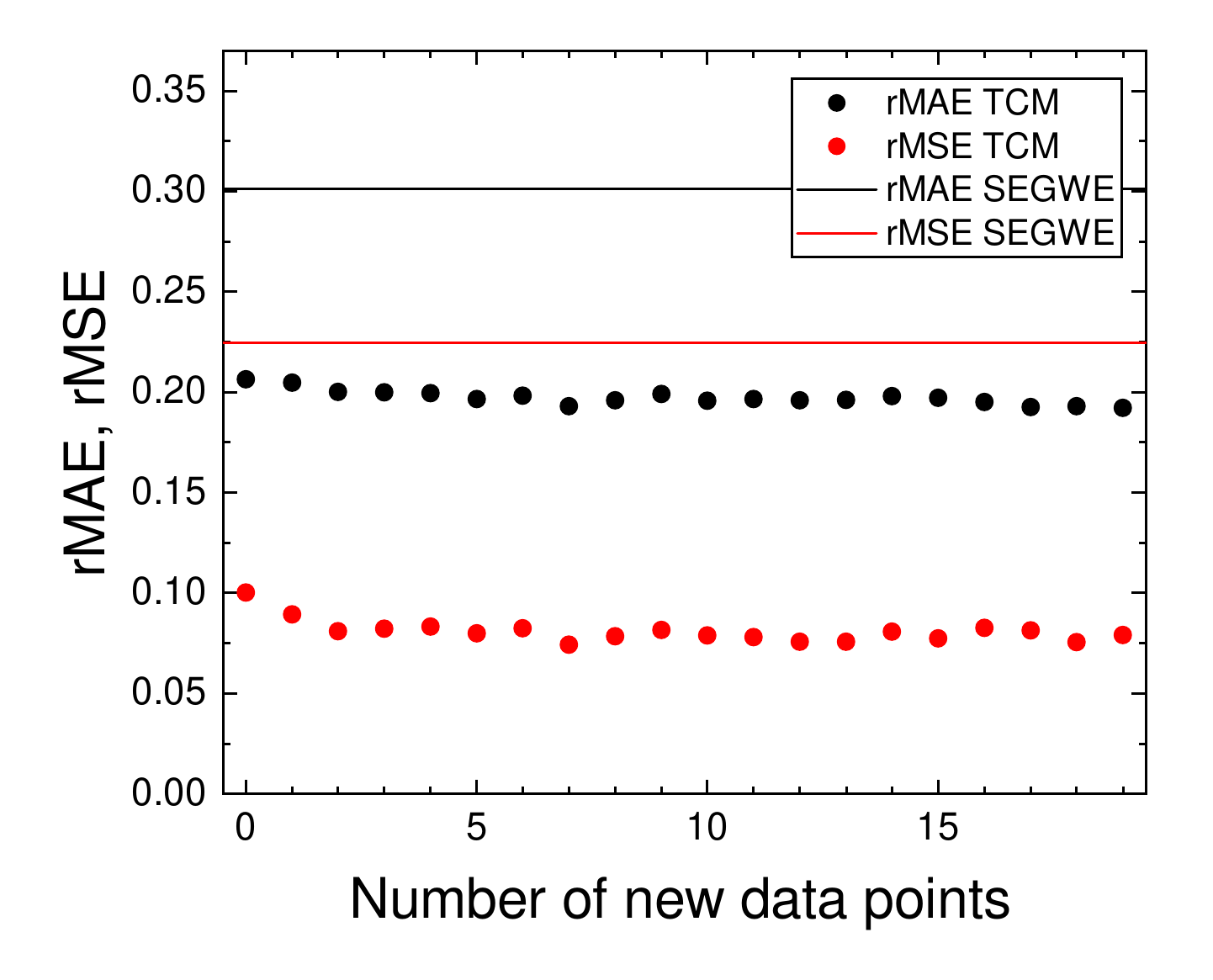}
    \end{subfigure}
    \begin{subfigure}{.49\textwidth}
    \centering
    \caption{333~K}
    \includegraphics[width=1\textwidth]{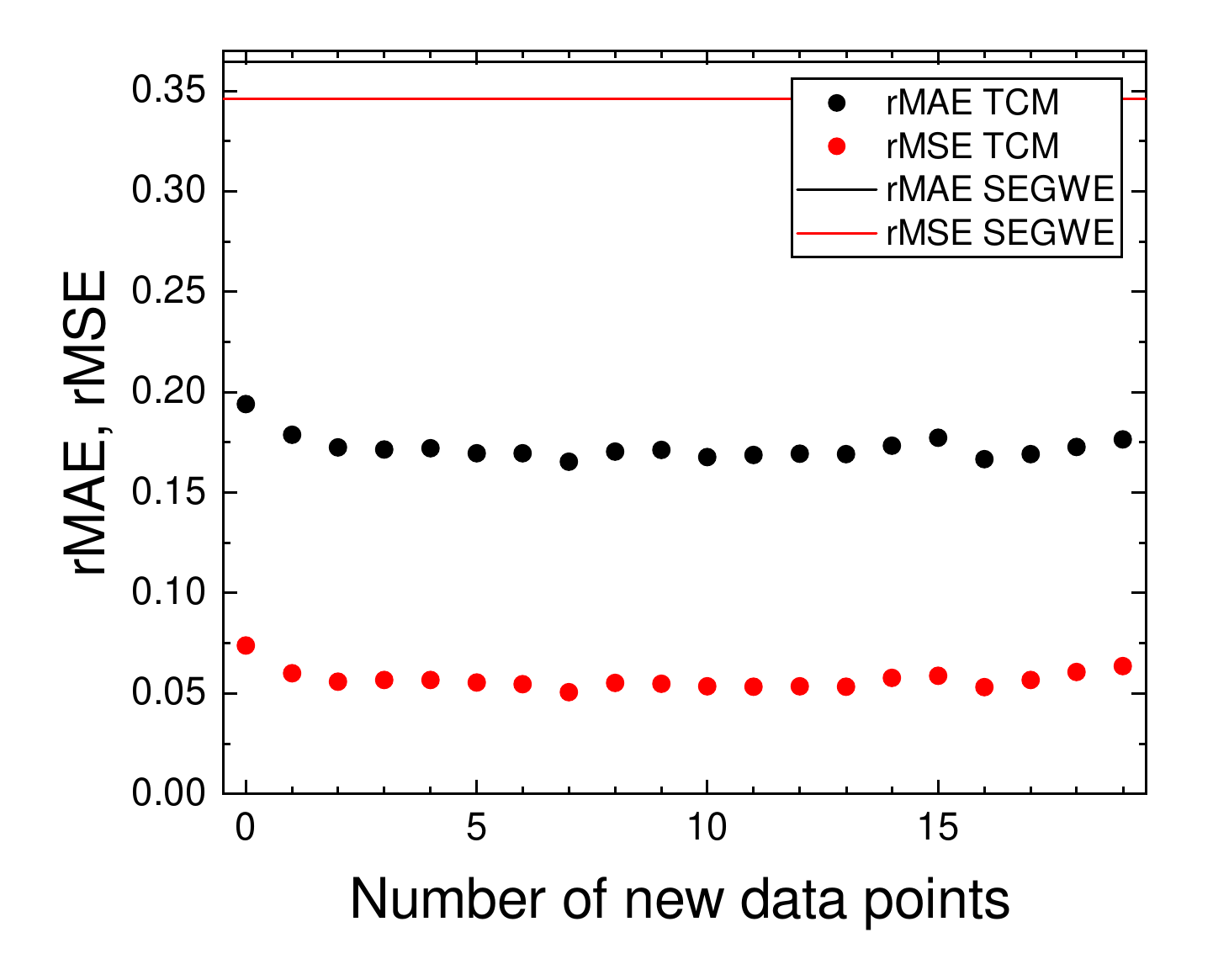}
    \end{subfigure}
    \caption{Performance of the TCM for predicting $D^{\infty}_{ij}$ in terms of relative mean absolute error (rMAE) and relative mean squared error (rMSE) as a function of the additional training data points selected using uncertainty sampling and measured in this work and comparison with SEGWE \cite{Evans2018}.}
    \label{fig:errors_all_SI}
\end{figure}

In Figure \ref{fig:errors_all_SI}, the influence of using the $D^{\infty,\mathrm{exp}}_{ij}(T)$ measured in this work in the training of the TCM on its prediction accuracy is shown. For this purpose, the values for rMAE and rMSE of the TCM evaluated by leave-one-out analysis based on the final available data set for all three temperatures is plotted as a function of the number of added training data points.

Figure \ref{fig:errors_all_SI} shows a slight decrease in the prediction errors of $D_{ijk}^\infty$. For 298~K, both rMAE and rMSE decrease over most periods, the effect for rMSE being more important, as might have been expected. However, for the two other temperatures, no significant decrease is observed during many episodes, leaving room for improvement in the query strategy. The comparison with the SEGWE results, which are also included in Figure \ref{fig:prediction_errors}, underline the superiority of the TCM method. 

\bibliography{achemso-demo}